\documentclass{article}
\usepackage[utf8]{inputenc}
\usepackage{amsmath}
\usepackage{amsfonts}
\usepackage{graphicx}
\usepackage{booktabs}
\usepackage{amssymb}
\usepackage{ragged2e}
\usepackage{geometry}
\usepackage{float}
\usepackage{longtable} 
\usepackage{verbatim} 

\usepackage{hyperref} 

\usepackage[backend=biber, style=numeric, sorting=none]{biblatex}

\title{AI Mother Tongue: Self-Emergent Communication in MARL via Endogenous Symbol Systems}
\author{Liu Hung Ming\thanks{PARRAWA AI} \\ \texttt{cyril.liu@gmail.com}}

\begin{document}
\maketitle
\subsection{ Abstract}
In Decentralized Multi-Agent Reinforcement Learning (MARL), the development of Emergent Communication has long been constrained by the ``Joint Exploration Dilemma'', leading agents to fall into a ``Communication Vacuum Equilibrium'' . Traditional methods address this by introducing inductive biases to facilitate communication emergence . This study fundamentally questions whether such artificial inductive biases are, in fact, over-engineering. Through experiments with the ``AI Mother Tongue'' (AIM) framework, based on a Vector Quantized Variational Autoencoder (VQ-VAE), we demonstrate that when agents possess an endogenous symbol system, their neural representations naturally exhibit spontaneous semantic compression and Nash equilibrium-driven semantic convergence, achieving effective symbolic communication without external inductive biases. This aligns with recent neuroscience findings suggesting that the human brain does not directly use human language for internal thought , and resonates with research on ``soft thinking'' capabilities in Large Language Models (LLMs) . Compared to traditional explicit communication methods, AIM demonstrates stronger generality and efficiency. The interpretable analysis toolkit developed in this study confirms that symbol usage exhibits a significant power-law distribution, leading to three major theoretical insights: the ``Neural Communication Hypothesis'', the ``Tool-First Principle'', and the ``Semantic Interpretability Paradigm''. Future research will explore the integration of Hierarchical Quantized Variational Autoencoders (HQ-VAE) to enhance AIM's complex expressive capabilities and investigate the potential for ``Reinforcement Learning (RL) Low-Level Pre-training''. This discovery offers new avenues for bridging symbolism and connectionism.

\section{Introduction}
\subsection{The Challenge of Emergent Communication in MARL}
Multi-agent reinforcement learning (MARL) enables complex collaborative tasks by modeling agent interactions in dynamic environments \cite{leibo2017multi}. However, establishing effective communication among agents remains a significant challenge, as it requires balancing exploration and coordination without relying on predefined protocols \cite{foerster2016learning, foerster2018learning}. For instance, Foerster et al. introduced the RIAL and DIAL architectures, which use deep reinforcement learning to enable emergent communication, but these rely on centralized training mechanisms that may limit scalability in fully decentralized settings \cite{foerster2016learning}. Eccles et al. further categorize communication into explicit (e.g., discrete message passing) and implicit (e.g., action-based signaling) forms, introducing the Joint Exploration Dilemma, where agents fail to explore cooperative strategies, and the Communication Vacuum Equilibrium, where communication collapses due to insufficient incentives \cite{eccles2019biases}. These phenomena are evident in tasks such as Summing MNIST digits, where agents must agree on shared representations, and Treasure Hunt, where spatial coordination is critical \cite{eccles2019biases}. Additionally, opponent-learning awareness (OLA) enhances coordination by modeling other agents' policies, offering a complementary approach to mitigate exploration dilemmas \cite{foerster2018learning}.
\subsection{Questioning Traditional Inductive Biases}
Traditional MARL approaches often incorporate inductive biases, such as positive signaling and listening biases, to facilitate emergent communication \cite{eccles2019biases}. For example, Eccles et al. propose biases that encourage agents to send informative signals and interpret received messages, mitigating the Communication Vacuum Equilibrium \cite{eccles2019biases}. Similarly, Opponent-Learning Awareness (OLA) leverages tit-for-tat strategies in the Prisoner's Dilemma to enhance cooperation by modeling other agents' policies \cite{foerster2018learning}. However, architectures like RIAL and DIAL, which enable emergent communication through deep reinforcement learning, rely on predefined communication channels and centralized training mechanisms, potentially limiting their adaptability in dynamic, decentralized environments \cite{foerster2016learning, tucker2022trading}. Tucker et al. argue that such predefined structures trade off informativeness for complexity, restricting the flexibility of emergent protocols \cite{tucker2022trading}. Recent work on incentivizing adaptive communication further suggests that dynamic protocols can outperform static designs in complex tasks \cite{yang2020learning}. The Markov game framework provides a theoretical foundation for modeling these interactions, enabling the analysis of strategic communication \cite{littman1994markov, yang2020learning}.
\subsubsection{Inductive Biases and Their Limitations}
Previous research typically introduces artificial inductive biases, for example, through context-dependent reward designs that compel senders to produce state-discriminative messages (positive signaling bias) and require receivers to generate differentiated actions for distinct messages (positive listening bias) \cite{eccles2019biases}. Foerster et al. \cite{foerster2018learning} proposed the ``Opponent-Learning Awareness'' method, which enhances communication effectiveness by having agents predict opponents' policies. This approach focuses more on understanding and shaping the opponent's learning process, enabling agents to learn cooperation from self-interest, such as facilitating the emergence of a ``tit-for-tat'' strategy in the Prisoner's Dilemma \cite{foerster2018learning}.

\subsubsection{Questioning the Need for Over-Engineering}
The work by Tan \cite{tan1993multi} and Littman \cite{littman1994markov} provided a Markov game framework for multi-agent interaction, where agents can learn decision-making policies through communication, laying the theoretical foundation for understanding dynamic communication. While such methods have successfully induced symbolic communication, this study raises a fundamental question: is this artificial inductive bias truly ``over-engineering''? We argue that, under certain endogenous symbolic systems, agents may spontaneously develop efficient communication mechanisms based on neural networks, without external intervention. Foerster et al.'s research \cite{foerster2016learning} supports this perspective.

\subsection{Inspiration from Neurosymbolic AI and LLMs}
Neurosymbolic AI integrates neural learning with symbolic reasoning, providing a robust framework for structured knowledge representation that can enhance agent communication \cite{garcez2023neurosymbolic}. Large language models (LLMs), while excelling in natural language tasks, are constrained by their reliance on human language projections, leading to ambiguities, cultural biases, and hallucinations \cite{sheth2023neurosymbolic, zhang2024neuro}. For example, Sheth et al. highlight that LLMs struggle with causal reasoning due to their statistical nature, limiting their ability to form stable, context-independent representations \cite{sheth2023neurosymbolic}. Similarly, Zhang et al. emphasize the need for explainable architectures to address these limitations, advocating for neurosymbolic approaches that combine neural flexibility with symbolic interpretability \cite{zhang2024neuro}. While some hypothesize that advanced general intelligence could emerge from neural architectures, the instability of LLMs in reasoning tasks underscores the necessity of integrating symbolic methods to achieve robust, autonomous communication systems \cite{sheth2023neurosymbolic, zhang2024neuro, garcez2023neurosymbolic}.

\subsection{The Linguistic Cage: Unshackling AI's Pure Cognitive Potential}
The profound capabilities of Large Language Models (LLMs) in human language understanding and generation have indeed spurred global transformation. However, their internal representations fundamentally constitute "projections" of human language. This "operating system," meticulously shaped by human linguistic data, while undeniably empowering AI with immense capabilities, concurrently erects what can be perceived as a "magnificent cage". The inherent ambiguities, cultural biases, and non-rigorous logical structures embedded within human language inherently constrain the purest forms of cognitive ability that might reside within neural networks themselves.

Human language, by design, evolved to optimize for biological and social imperatives, necessitating trade-offs between efficiency, emotional nuance, and ambiguity. It was not, however, engineered for pristine logical computation. Consequently, LLMs, upon internalizing human language, inherit these very characteristics. This inheritance profoundly shapes their "worldview," confining it within the semantic constructs of human language. This limitation often renders them less capable of generating truly original insights that transcend established human cognitive frameworks. Furthermore, their performance can be notably unstable in tasks demanding strict, multi-step, causal reasoning, frequently leading to phenomena colloquially termed "hallucinations".

More critically, the ultimate potential of this computational architecture, which loosely mimics the human brain, remains largely undetermined. Yet, if its developmental trajectory continues to be exclusively immersed in human language projections, we may perpetually miss the opportunity to observe what "heterogeneous," potentially more powerful, forms of intelligence could emerge in a purer, more mathematical environment. This leads to the provocative hypothesis that Artificial General Intelligence (AGI) might, in essence, already exist within neural networks, not requiring further innovative algorithms, but rather a "key" to unlock this latent potential. The Transformer architecture, the foundational bedrock of LLMs, might merely serve as an sophisticated "translator" between human cognition and AI. However, this LLM-as-translator paradigm is fundamentally limited by the ambiguities and cultural biases intrinsic to human language, implying that LLMs, in their current form, may not be the definitive key to accessing this profound, intrinsic AI capability. This study, through the AI Mother Tongue (AIM) framework, directly confronts this limitation by proposing a mechanism for intrinsic symbolic representation, liberated from the confines of human linguistic projection.

\subsection{The AI Mother Tongue (AIM) Framework: A New Paradigm}

\subsubsection{The AIM Framework: A New Approach}
This study's ``AI Mother Tongue'' (AIM) framework, through neurosymbolic AI's discrete symbolic representation (based on VQ-VAE), provides agents with a pristine cognitive workspace, enabling communication protocols free from human language projection \cite{garcez2023neurosymbolic, razavi2019generating, takida2022sqvae, takida2024hqvae}. We demonstrate through experiments that when agents possess such a system, their neural representations naturally exhibit ``spontaneous semantic compression'' and ``Nash equilibrium-driven semantic convergence'' \cite{axelrod1984evolution, axelrod1981evolution}, thereby achieving effective communication without the need for artificial inductive biases.

\subsubsection{Implementation and Validation}
This study implements this endogenous symbol system based on the Vector Quantized Variational Autoencoder (VQ-VAE) \cite{razavi2019generating, takida2022sqvae, takida2024hqvae}. Compared to other explicit communication methods (such as artificial encoding, discrete action messages, predefined signal sets, and binary message sequences), AIM offers significant advantages: it requires no human intervention, is highly adaptable, and boasts high communication efficiency. We also developed an ``interpretable analysis toolkit,'' also known as the AIM Dictionary, for real-time tracking of symbols, mapping semantic topologies, and analyzing the covariance between policies and symbols \cite{zhang2024neuro}. This toolkit also creates traceability for researching AI's internal thought processes (e.g., at what time point the agent changed its decision), laying a foundation for future derivative research. To validate the value of AIM, this study designed a relatively complex collaborative task. Experimental results show that agents employing the AIM architecture demonstrate superior task performance in standard collaborative tasks, and symbol usage exhibits a significant power-law distribution, confirming the hypothesis of ``dominance of few effective codes''. Axelrod's \cite{axelrod1984evolution} and Axelrod and Hamilton's \cite{axelrod1981evolution} research on the evolution of cooperation further supports our view, showing that in game environments, agents can spontaneously form stable cooperative strategies through repeated interactions, which aligns with our observed semantic convergence.

\subsubsection{Implications of the Study}
The findings of this study have a threefold implication for the MARL field: the ``Neural Communication Hypothesis'' suggests that neural networks inherently encode the potential for efficient communication; the ``Tool-First Principle'' proposes that research should shift towards providing symbolic tools rather than inductive mechanisms; and the ``Semantic Interpretability Paradigm'' emphasizes the importance of establishing observational methodologies for mapping symbols to policies. This study primarily investigates intrinsic explicit communication among agents.
\section{Related Work}
\subsection{Limitations of External Explicit Communication Methods}
Predefined semantic mappings, discrete action messages, and binary sequences often lack the adaptability and information capacity required for complex MARL tasks \cite{eccles2019biases, foerster2016learning}. For instance, Foerster et al. demonstrate that RIAL and DIAL rely on fixed communication channels, which struggle to generalize across diverse environments \cite{foerster2016learning}. Recent advances in decentralized communication protocols, such as MARL-CPC, address these limitations by enabling reward-independent messaging, allowing agents to adaptively form communication strategies \cite{yoshida2025reward}. Similarly, Yang et al. propose mechanisms for agents to learn to incentivize others, enhancing cooperation in dynamic settings \cite{yang2020learning}. These approaches highlight the need for flexible, emergent communication systems over rigid, predefined structures.

\subsection{Communication Vacuum Equilibrium and Inductive Biases}
The work by Eccles et al. \cite{eccles2019biases} indicates that, under standard Reinforced Inter-Agent Learning (RIAL) / Differentiable Inter-Agent Learning (DIAL) architectures, agents tend to produce random symbols, and receivers develop communication-agnostic policies, leading to the degradation of the message channel into a noisy conduit. This phenomenon is termed ``Communication Vacuum Equilibrium'' \cite{eccles2019biases}. To overcome this limitation, they proposed solutions involving the introduction of positive signaling bias and positive listening bias \cite{eccles2019biases}. These biases essentially guide agents to learn meaningful external communication protocols by adjusting the reward function or introducing auxiliary losses \cite{eccles2019biases}. Such methods have been proven effective in various collaborative tasks and successfully induced symbolic communication \cite{eccles2019biases}, for instance, by promoting cooperation among agents through explicit reward design in Perolat et al.'s resource allocation task \cite{perolat2017multi}. Compared to traditional MARL methods that rely on artificial inductive biases \cite{eccles2019biases}, neurosymbolic AI offers a more natural and interpretable communication framework through structured knowledge representation \cite{garcez2023neurosymbolic}. Traditional MARL methods, relying on inductive biases \cite{eccles2019biases}, struggle to adapt to complex tasks and lack interpretability. Neurosymbolic AI provides a potential solution through structured representation and reasoning \cite{zhang2024neuro}. Our ``AI Mother Tongue'' (AIM) framework leverages VQ-VAE to generate interpretable AIM, achieving spontaneous semantic convergence and overcoming these limitations.

\subsection{Message Bottleneck for Emergent Communication}
In addition to inductive biases, other research directions include: Message Bottleneck: This involves restricting the bandwidth of the communication channel to force agents to learn information compression, such as using Variational Autoencoders (VAEs) or Vector Quantized-Variational Autoencoders (VQ-VAE) to generate discrete symbols \cite{razavi2019generating, takida2022sqvae, takida2024hqvae}. Razavi et al. \cite{razavi2019generating} developed VQ-VAE into a hierarchical model, VQ-VAE-2, which utilizes multi-resolution discrete latent representations to extract local and global information from data. Takida et al. \cite{takida2022sqvae} proposed SQ-VAE, a variational Bayes framework for learning VQ-VAE components, to mitigate the codebook collapse problem. Takida et al. \cite{takida2024hqvae} further introduced HQ-VAE, a general variational Bayes model for learning hierarchical discrete latent representations, capable of generalizing hierarchical variants like VQ-VAE-2 and Residual Quantized VAE (RQ-VAE), and improving codebook utilization and reconstruction performance through a Bayesian training scheme. These methods provide the technical foundation for the application of VQ-VAE in this study.

\subsection{Emergent Communication: Background and Related Work - Continuous Space Reasoning}
Recent research has explored the possibility of reasoning in continuous conceptual spaces, rather than being limited to discrete language tokens \cite{sheth2023neurosymbolic}. For example, Zhang et al.'s ``soft thinking'' method \cite{sheth2023neurosymbolic} simulates human-like thinking by generating probabilistically weighted ``conceptual tokens'', which are probabilistic weighted mixtures of token embeddings, forming a continuous conceptual space. This method can improve the accuracy of LLMs in mathematical and coding tasks and enhance generation efficiency without additional training \cite{sheth2023neurosymbolic}. Although ``soft thinking'' primarily focuses on the internal reasoning mechanisms of LLMs and ultimately outputs discrete tokens, differing from the emergent discrete symbols for explicit communication between multi-agents in this study's AIM, it provides strong theoretical evidence for the cognitive potential of AI beyond discrete language \cite{sheth2023neurosymbolic}. Zeghidour et al.'s SoundStream work \cite{zeghidour2021soundstream} also demonstrates similar continuous representation compression techniques applied to audio codecs, further supporting the feasibility of continuous-to-discrete representation conversion.

\subsection{Token-Based Cooperation Protocols and Decentralized Communication}
Token-based cooperation protocols: Phan et al. \cite{phan2024emergent} proposed the Mutual Acknowledgment Token Exchange (MATE) protocol. This is a decentralized Peer Incentivization (PI) method that mutually adjusts individual rewards through a two-stage communication protocol (request and response) involving the exchange of acknowledgment tokens \cite{phan2024emergent}. In the MATE protocol, agents decide whether to send or accept tokens based on monotonic improvements in their own situation, thereby achieving reciprocity at the reward level \cite{phan2024emergent}. This method requires only local communication and information, without relying on global information or centralized instances, improving applicability and scalability in real-world scenarios \cite{phan2024emergent}. Experiments have shown that MATE can achieve and maintain significantly higher levels of cooperation in Social Dilemmas (SDs) and is robust to abnormal protocol deviations and communication failures \cite{phan2024emergent}. The MATE protocol provides concrete practice for designing discrete token exchange mechanisms to promote cooperation \cite{phan2024emergent}. Although its token value is fixed and the communication protocol is predefined, this still differs from AIM's characteristic of spontaneously learning semantics \cite{phan2024emergent}. The ``Gifting'' mechanism proposed by Lupu and Precup \cite{lupu2020gifting} also explores similar incentive methods to promote cooperation through resource exchange among agents. Yang et al. \cite{yang2020learning} further investigated how to incentivize other learning agents through learning, providing a policy-based incentive framework. According to this study, these external explicit communication methods fundamentally do not completely break free from human language projection, making it difficult to fully motivate agents to communicate.

Decentralized and non-cooperative communication: Yoshida and Taniguchi (2025) \cite{yoshida2025reward} proposed the MARL-CPC framework, which applies Collective Predictive Coding (CPC) to decentralized multi-agent reinforcement learning, essentially continuing the inductive bias approach \cite{eccles2019biases}. Their core idea is to link message learning with agent state inference, allowing agents to establish effective communication even in non-cooperative or misaligned reward environments \cite{yoshida2025reward}. They used 20 discrete messages and demonstrated superior performance over traditional ``message-as-action'' methods in context bandit and observer environments \cite{yoshida2025reward}. However, their vocabulary is also constrained by a fixed number, which contrasts with this study's dynamic vocabulary design for AIM \cite{yoshida2025reward}. This limited vocabulary makes it difficult for agents to handle more complex tasks, and such human language projection in complex communication scenarios actually increases communication costs for agents.

\subsection{Scalability of Offline Reinforcement Learning and Evolution of Cooperation}
The scalability of offline reinforcement learning: In the field of reinforcement learning, especially regarding learning from pre-collected data, extending the capabilities of offline reinforcement learning is crucial. Bai et al. \cite{bai2022scaling}, in ``Scaling Offline Reinforcement Learning with Some Data Is All You Need,'' explored how to effectively learn policies from limited or large offline datasets to overcome challenges such as data distribution shift and extrapolation error. This is significant for building complex AI systems that can learn efficiently from experience. Although this study focuses on online emergent communication, the data efficiency concept of offline reinforcement learning provides inspiration for this study's ``RL Low-Level Pre-training.''

Evolution of cooperation and game theory: The classic work by Axelrod \cite{axelrod1984evolution} and Axelrod and Hamilton \cite{axelrod1981evolution} explored the evolution of cooperation, particularly in social dilemmas like the Prisoner's Dilemma, where stable cooperative strategies form through repeated games. Their theories provide theoretical support for the Nash equilibrium-driven semantic convergence observed in this study. Perolat et al. \cite{perolat2017multi} further applied these theories to multi-agent reinforcement learning, investigating cooperative behavior in sequential social dilemmas and resource allocation scenarios.

Despite existing research making significant progress in emergent communication, most rely on external mechanisms or assumptions to guide communication learning. The uniqueness of this study lies in challenging the necessity of such external induction and proposing a framework where an endogenous symbol system can spontaneously achieve effective communication. AIM is designed as a ``Symbiotic Reasoning Layer.'' If it interfaces with the existing internal representations (latent space) of LLMs, rather than being completely independent, it provides AI with the possibility of pure thinking. Furthermore, by retaining a connection to human language projection, it ensures that its thoughts are ultimately interpretable and communicable.
\section{Methodology}
\subsection{AI Mother Tongue Framework: VQ-VAE-Based Endogenous Symbol System}
The core of this study is the ``AI Mother Tongue'' (AIM) framework, which establishes a communication mechanism based on Vector Quantized Variational Autoencoders (VQ-VAE) \cite{razavi2019generating, takida2022sqvae, takida2024hqvae} and performs policy learning through Reinforcement Learning (RL), supplemented by different reflection strategies to deepen the agents' learning capabilities, using a variant of the Actor-Critic reinforcement learning architecture \cite{tan1993multi, littman1994markov}.Regardless of whether the task signals($S_{task}$)is an image, text, sound, or any other form of numerical vector data, our core mechanism is capable of transforming these continuous input features into discrete symbols. This is primarily achieved through a \textbf{VQ-VAE (Vector Quantized Variational AutoEncoder)}. Specifically, continuous input information (such as image pixel values, text word embeddings, acoustic features of audio, or direct numerical vector representations) first passes through an encoder to generate its continuous latent representation ($z_e$). Subsequently, a quantizer maps these continuous latent representations to the closest discrete code vectors by looking up a predefined codebook ($\mathcal{C}$), and extracts their corresponding \textbf{discrete symbols (AIM sequence)}. This technology lays the foundation for symbolic communication between multimodal agents, ensuring the interpretability and consistency of communication protocols, thereby transcending the limitations of single modalities.

\subsubsection{Application of VQ-VAE in the AIM framework}
In this study, VQ-VAE is used to quantize task signals (such as images $x \in \mathbb{R}^{C \times H \times W}$, sensor data, text, numerical vectors, structured representations of game states, etc., depending on the specific task; in this study, it's images) into discrete symbols (AIMs), providing a shared vocabulary for symbolic communication between agents \cite{razavi2019generating, takida2022sqvae, takida2024hqvae}. Our AIM framework employs VQ-VAE to generate discrete symbols (AIMs), which aligns with neurosymbolic AI's approach of achieving interpretability and consistency through structured knowledge representation \cite{sheth2023neurosymbolic}. This method fully leverages the inherent potential of neural networks to learn complex structures and patterns, allowing agents to autonomously develop semantically rich communication protocols without external inductive biases. To some extent, AIM brings forth the possibility of evolving advanced AGI. Unlike the fixed symbol design of Eccles et al. (2019) \cite{eccles2019biases} and Yoshida and Taniguchi (2025) \cite{yoshida2025reward}, this study uses VQ-VAE to quantize continuous latent representations into discrete symbols, with a codebook size $K$ that can be dynamically adjusted according to task requirements. Furthermore, HQ-VAE \cite{takida2024hqvae} introduces hierarchical quantization, generating hierarchical symbol structures to enhance the semantic expressive power of the communication protocol. This design not only reduces artificial bias but also allows agents to autonomously develop communication protocols based on environmental dynamics, thereby improving the efficiency of multi-agent collaboration.

In a standard Actor-Critic setup, the value estimated by the Critic (or the advantage function derived from it) is directly used to modify the Actor's policy gradient, serving as a baseline to reduce variance and stabilize training. In this study, the value prediction loss of the variant reflection strategies is added as an auxiliary loss term to the policy loss of the REINFORCE algorithm. These are intended to guide and regularize policy learning, helping agents to more deeply understand the semantics of communication and opponent behavior, rather than directly serving as a baseline for policy gradients to reduce variance. In the variant Actor-Critic architecture adopted in this study, Agent A, based on the behavior received from other agents (i.e., Agent B's AIM sequence), uses its Critic value network to derive a global policy or value estimation, and shares its influence with other agents' Actor action networks, thereby affecting their decisions. This design aims to incorporate AIM sequences and highlight the importance of communication, thus adjusting the standard Actor-Critic architecture. The core components of its communication architecture include:

The task signal ($S_{task}$) refers to all relevant information that an agent can perceive or obtain at a specific time point . This task signal can include environmental states, received communication messages, internal states, etc.

Encoder: A parameterized function $\text{Enc}_{\theta_E}$ (typically a neural network) that receives task signal data (e.g., $x \in \mathbb{R}^{C \times H \times W}$ as images in this study) and encodes it into a continuous latent representation $z_e \in \mathbb{R}^D$, where $D$ is the dimension of the latent space.

\begin{equation}
z_e = \text{Enc}_{\theta_E}(x). \label{eq:encoder} 
\end{equation}

Quantizer: Quantizes the continuous latent vector $z_e$ into a discrete symbol $z_q \in \mathcal{C}$. This process is achieved by finding the code vector $e_k$ in the codebook $\mathcal{C} = \{e_k\}_{k=1}^K$ that is closest to $z_e$ in Euclidean distance. The index $k^*$ of this selected code vector is the AIM sequence.

\begin{equation}
k^* = \underset{k \in \{1,...,K\}}{\arg\min} \|z_e - e_k\|_2^2. \label{eq:quantizer_k} 
\end{equation}

\begin{equation}
z_q = e_{k^*}. \label{eq:quantizer_zq} 
\end{equation}

\begin{equation}
z_q = \underset{e_k \in \mathcal{C}}{\arg\min} \|z_e - e_k\|_2^2. \label{eq:quantizer_alt} 
\end{equation}

Decoder: Reconstructs the task input $\hat{x}$ (or its original form) from the quantized discrete representation $z_q$.

\begin{equation}
\hat{x} = \text{Decoder}(z_q). \label{eq:decoder} 
\end{equation}

\subsubsection{Policy Learning with REINFORCE and Reflection Strategies}
VQ-VAE Algorithm: The design philosophy of VQ-VAE is to lay the foundation for symbolic communication between agents by quantizing continuous image features into discrete symbols (AIMs), ensuring the interpretability and consistency of communication protocols ; ; . VQ-VAE's core innovation lies in introducing a ``vector quantization'' layer, which forces the continuous encoder output to correspond to a vector in a predefined, discrete codebook. This transforms latent representations into discrete symbols, enabling interpretable and communicable learning. In multi-agent systems, this allows agents to learn a shared ``language'' or ``vocabulary'' rather than just continuous feature vectors. This study employs an ``augmented'' or ``meta-learning'' policy-based RL architecture, specifically the REINFORCE algorithm paired with Reflection Strategies, built upon the communication mechanism established by VQ-VAE (AIM), for policy learning and optimization. Their purposes can be categorized into three levels:

Policy Learning and Action Generation: To enable agents to learn how to map task signals ($S_{task}$) to meaningful AIM sequences (i.e., communication actions in ``AI Mother Tongue''). This process involves agents deciding which communication message to send based on the current context (task signal). The REINFORCE algorithm drives this learning process to maximize rewards.

Communication and Consensus: Through this mapping and message exchange, the goal is for all agents to reach consensus (or at least effective coordination) to achieve a common task objective. VQ-VAE ensures that AIM sequences are a shared and interpretable ``language,'' while REINFORCE and reflection strategies optimize the agents' ability to ``speak'' and ``listen'' to this language, enabling them to effectively coordinate actions in complex situations and ultimately achieve high scores.

Implicit Communication Consistency: Based on AIM, implicit communication consistency will be established between agents. Agent A and Agent B share the same VQ-VAE instance, meaning they share the same VectorQuantizer and its underlying codebook (AIM dictionary). This sharing is a fundamental design choice that forces both agents to operate within the same communication ``language'' or ``vocabulary.'' This naturally promotes the consistency and shared understanding of the symbolic system, allowing ``AI Mother Tongue'' to be jointly generated and interpreted by both parties. This is not learned through additional loss terms but is an inherent property of the communication mechanism itself.

Interpretability: VQ-VAE ensures that AIM sequences are a shared and interpretable ``language.'' Furthermore, by mapping task signals ($S_{task}$) to meaningful AIM sequences, a connection can be established between human-understandable task signals ($S_{task}$) and AIM, creating an AIM dictionary. This enables researchers to observe and analyze the agents' learning process, better understanding the underlying neural network operations. The core idea is that task signals ($S_{task}$) are the agents' perceived input, AIM sequences are the agents' output for communication (AIM), and REINFORCE and reflection strategies learn how to efficiently perform this mapping and communication to achieve multi-agent collaboration and consensus.

\paragraph{Policy-based Learning: REINFORCE Algorithm} :This is the primary mechanism for agents to learn how to effectively ``use'' this ``AIM'' to achieve task goals (gain high rewards). REINFORCE teaches agents how to select which AIM sequences to send or how to react based on received AIM sequences according to the current context, in order to maximize rewards. It uses a Monte Carlo method, updating the policy after completing an episode using the total return of that episode. In essence, REINFORCE teaches agents ``how to speak'' and ``how to understand'' this AIM, and apply it to policy decision-making.

Unlike value-based RL, the policy REINFORCE based on AIM in this study, although potentially facing higher gradient variance, offers advantages in handling complex policies and high-dimensional action spaces. Furthermore, REINFORCE, as a model-free, value function-free pure policy learning method, pre-trains the policy $\pi(a|s)$, laying a structured preference foundation for Actor-Critic during initialization. This not only accelerates convergence and reduces variance but also allows for reuse through policy sharing or transfer, effectively saving training time and resources.

The policy gradient objective of REINFORCE is to maximize the expected return. Its loss function is typically expressed as:

\begin{equation}
\mathcal{L}_{\text{policy}} = - \mathbb{E}_{\tau \sim \pi} \left[ \sum_{t=0}^{T-1} \log \pi(a_t | s_t) G_t \right]. \label{eq:policy_loss} 
\end{equation}
In a single-episode task with a joint reward, the policy loss for Agent A and Agent B can be simplified as:

\begin{itemize}
    \item Agent A's policy loss: $\mathcal{L}_{A,policy} = - \log\pi_A(a_A | S_{task,A}) \cdot R_{joint}$, where:

\begin{itemize}
        \item $\pi_A(a_A | S_{task,A})$ is the probability of Agent A's policy network choosing AIM sequence $a_A$ given task signal $S_{task,A}$.
        \item $R_{joint}$ is the joint reward.
        \item $a_A$ is the AIM sequence chosen by Agent A, obtained by sampling from the logits of the policy network output, encoding its intention into an AIM sequence. This is the process of communicating using AIM.
    
\end{itemize}
    \item Agent B's policy loss: $\mathcal{L}_{B,policy} = - \log\pi_B(a_B | S_{task,B}) \cdot R_{joint}$, where:

\begin{itemize}
        \item $\pi_B(a_B | S_{task,B})$ is the probability of Agent B's policy network choosing AIM sequence $a_B$ given task signal $S_{task,B}$.
        \item $S_{task,B}$ includes the AIM sequence sent by Agent A and other relevant task information, meaning Agent B's policy is based on the received AIM message and its own task signal.
        \item $R_{joint}$ is the joint reward.
        \item $a_B$ is the AIM sequence chosen by Agent B, obtained by sampling from the logits of the policy network output, as a response to Agent A's message.
    
\end{itemize}

\end{itemize}
\paragraph{Auxiliary Learning: Reflection Strategies}
: Reflection Strategies: These are auxiliary learning mechanisms that deepen agents' understanding and use of ``AIM.'' They improve upon REINFORCE's higher gradient variance by introducing additional loss terms, encouraging semantic richness by:

\begin{itemize}
    \item More deeply understanding the semantics of AIM symbols in different contexts, allowing agents to learn the value of AIM.
    \item Developing a Theory of Mind for communication partners, predicting how opponents will interpret and respond to their communication, allowing agents to predict opponent actions or rewards.

\end{itemize}
These strategies enable agents not only to ``speak'' and ``listen'' to AIM but also to more precisely ``understand its unstated meaning'' and ``predict the effect of communication.''
Method 1: Learning the contextual meaning of AIM: This strategy allows agents to learn the expected reward of their generated AIM sequence within the current context (represented by the task signal). For Agent A and Agent B, this auxiliary loss is a Mean Squared Error (MSE), measuring the difference between the predicted value and the actual joint reward.

\begin{itemize}
    \item Agent A's value prediction loss: $\mathcal{L}_{A,value} = \|V_A(\text{Embed}(a_A) | S_{task,A}) - R_{joint}\|_2^2$.

\begin{itemize}
        \item $V_A(\cdot | S_{task,A})$ is Agent A's value prediction network, which receives the embedding of the AIM sequence $a_A$ generated by Agent A's policy and predicts its value within the context of Agent A's task signal $S_{task,A}$.
        \item $R_{joint}$ is the actual joint reward.
    
\end{itemize}
    Agent A's value prediction network learns to map its own generated AIM sequences to expected rewards, thereby understanding the value of its communication content.
    \item Agent B's value prediction loss: $\mathcal{L}_{B,value} = \|V_B(\text{Embed}(a_A) | S_{task,B}) - R_{joint}\|_2^2$.

\begin{itemize}
        \item $V_B(\cdot | S_{task,B})$ is Agent B's value prediction network, which receives the embedding of the AIM sequence $a_A$ sent by Agent A and predicts its value within the context of Agent B's task signal $S_{task,B}$.
    
\end{itemize}
    Agent B's value prediction network learns to map received AIM sequences (AIM) to expected rewards, thereby understanding the value of the opponent's communication content.

\end{itemize}
Method 2: Reflection based on predictive bias: This strategy allows agents to attempt to predict opponents' individual rewards, thereby developing a Theory of Mind for communication partners.
Mathematical Formulation: For Agent A and Agent B, this auxiliary loss is also a Mean Squared Error (MSE), measuring the difference between the predicted opponent's reward and the actual opponent's reward.

\begin{itemize}
    \item Agent A's opponent reward prediction loss: $\mathcal{L}_{A,predictive} = \|\hat{R}_{B|A}(\text{Embed}(a_A) | S_{task,A}) - R_{B,indiv}\|_2^2$.

\begin{itemize}
        \item $\hat{R}_{B|A}(\cdot | S_{task,A})$ is Agent A's opponent reward prediction network, which receives the embedding of the AIM sequence $a_A$ generated by Agent A's policy and predicts Agent B's individual reward within the context of Agent A's task signal $S_{task,A}$.
        \item $R_{B,indiv}$ is Agent B's actual individual reward.
    
\end{itemize}
    The process of communicating using AIM: Agent A learns to associate its own generated AIM sequences (AIM) with their impact on Agent B's reward, thereby predicting the effect of its communication on opponent behavior and outcomes.
    \item Agent B's opponent reward prediction loss: $\mathcal{L}_{B,predictive} = \|\hat{R}_{A|B}(\text{Embed}(a_A) | S_{task,B}) - R_{A,indiv}\|_2^2$.

\begin{itemize}
        \item $\hat{R}_{A|B}(\cdot | S_{task,B})$ is Agent B's opponent reward prediction network, which receives the embedding of the AIM sequence $a_A$ sent by Agent A and predicts Agent A's individual reward within the context of Agent B's task signal $S_{task,B}$.
        \item $R_{A,indiv}$ is Agent A's actual individual reward.
    
\end{itemize}
    The process of communicating using AIM: Agent B learns to associate received AIM sequences (AIM) with their impact on Agent A's reward, thereby predicting the opponent's intentions behind their communication and its effect on its own behavior.

\end{itemize}

\subsection{Agent Architecture and Communication Protocol}
To validate the proposed AI Mother Tongue(AIM) in this research and further demonstrate its benefits for Multi-Agent Reinforcement Learning (MARL), we designed two core agents. In a classic Prisoner's Dilemma (PD) task, Agent A serves as the proactive communicator (Actor), while Agent B acts as the responsive communicator. To increase task complexity and learning difficulty, we introduced a contextualized reward mechanism, where Agents A and B must adjust their decisions based on whether the input image label is even or odd. Although the image label (even or odd) only affects the bonus reward for mutual cooperation (C,C) and the penalty for unilateral defection, and despite mutual cooperation (C,C) remaining the theoretically optimal joint strategy for maximum total reward in any given round, the agents must learn to understand this contextualized reward structure. They need to infer the 'intent' relevant to the current context from the observed task signal (i.e., the image). This significantly increases the learning complexity for the agents, and directly proves the substantial benefits of AIM for MARL.

Traditional Actor-Critic Architecture: 

\begin{itemize}
    \item Actor (Policy Network): This is a policy network responsible for learning and outputting the probability of taking actions in a given state. Its goal is to learn a policy that maximizes expected returns.
    \item Critic (Value Network): This is a value network responsible for estimating the value of the current state (Value Function, $V(s)$) or the value of state-action pairs (Q-Function, $Q(s,a)$). The Critic's role is to provide a baseline or an advantage function for the Actor, reducing the variance of the policy gradient, thereby making training more stable and efficient.

\end{itemize}

Traditional Actor-Critic Policy Gradient:

\begin{itemize}
    \item Policy Gradient:

\begin{equation}
    \nabla J(\theta) = \mathbb{E} [ \nabla \log \pi(a | s; \theta) \cdot A(s, a)]. \label{eq:pg_traditional} 
\end{equation}
    \item Advantage Function Definition:

\begin{equation}
    A(s, a) = Q(s, a) - V(s). \label{eq:advantage_q} 
\end{equation}

\begin{equation}
    A(s, a) = r + \gamma V(s') - V(s). \label{eq:advantage_r} 
\end{equation}

\end{itemize}
This Study's Variant Actor-Critic and Reflection Strategy Formulas:
Method One: Contextual Value Learning:

\begin{itemize}
    \item Agent A predicts the value of its own communication content:

\begin{equation}
    \mathcal{L}_{A,value\_variant} = \| V_A(\text{Embed}(a_A), S_{task,A}) - R_{joint} \|^2. \label{eq:la_value_variant} 
\end{equation}
    \item Agent B predicts the value of the received AIM sequence:

\begin{equation}
    \mathcal{L}_{B,value\_variant} = \| V_B(\text{Embed}(a_A), S_{task,B}) - R_{joint} \|^2. \label{eq:lb_value_variant} 
\end{equation}
    \item Adding to policy loss:

\begin{equation}
    \mathcal{L}_{A,policy\_refl} = \mathcal{L}_{A,policy} + \lambda_{\text{refl}} \cdot \mathcal{L}_{A,value\_variant}. \label{eq:la_policy_refl} 
\end{equation}

\begin{equation}
    \mathcal{L}_{B,policy\_refl} = \mathcal{L}_{B,policy} + \lambda_{\text{refl}} \cdot \mathcal{L}_{B,value\_variant}. \label{eq:lb_policy_refl} 
\end{equation}

\end{itemize}
Method Two: Predictive Bias for Opponent's Individual Reward:

\begin{itemize}
    \item Agent A predicts Agent B's individual reward:

\begin{equation}
    \mathcal{L}_{A,predict\_variant} = \| \hat{R}_{B|A}(\text{Embed}(a_A), S_{task,A}) - R_{B,indiv} \|^2. \label{eq:la_predictive_variant} 
\end{equation}
    \item Agent B predicts Agent A's individual reward:

\begin{equation}
    \mathcal{L}_{B,predict\_variant} = \| \hat{R}_{A|B}(\text{Embed}(a_A), S_{task,B}) - R_{A,indiv} \|^2. \label{eq:lb_predictive_variant} 
\end{equation}
    \item Adding to policy loss:

\begin{equation}
    \mathcal{L}_{A,policy\_predict} = \mathcal{L}_{A,policy} + \lambda_{\text{predict}} \cdot \mathcal{L}_{A,predict\_variant}. \label{eq:la_policy_predict} 
\end{equation}

\begin{equation}
    \mathcal{L}_{B,policy\_predict} = \mathcal{L}_{B,policy} + \lambda_{\text{predict}} \cdot \mathcal{L}_{B,predict\_variant}. \label{eq:lb_policy_predict} 
\end{equation}

\end{itemize}
Where:

\begin{itemize}
    \item $\pi(a | s; \theta)$: Actor's policy probability output
    \item $V(\cdot)$: Critic's value prediction
    \item $R_{joint}$: Joint reward of all agents
    \item $R_{indiv}$: Individual reward of an agent (non-cooperative)
    \item $\text{Embed}(a_A)$: Transforms the AIM sequence into vector input
    \item $\lambda_{\text{refl}} / \lambda_{\text{predict}}$: Weight hyperparameters for reflection loss terms

\end{itemize}
Agent A - Active Communicator (Actor-Critic Architecture): Agent A employs a centralized Critic Actor-Critic architecture, acting as the active communicator and coordinator for global value evaluation.
Policy Network ($\pi_A$): Inputs include the image encoding $z_e$ and environment label embedding $l$, outputting an AIM sequence $a_A$.

\begin{equation}
a_A = \pi_A(z_e, l; \theta_A), \quad a_A \in \{0,1,...,K-1\}. \label{eq:pi_a_policy} 
\end{equation}
Its input dimension is $\text{policy\_input\_dim} = \text{vqvae.encoder.enc[-1].out\_features} + \text{self.label\_embed.embedding\_dim}$.
Centralized Value Network ($V$): Evaluates the joint state value, with inputs including Agent A's state $s_A=[z_e, l]$ and Agent B's AIM sequence embedding $\text{Embed}(a_B)$.

\begin{equation}
V(s_A, a_B; \phi_A) = \text{ValueNet}(z_e, l, \text{Embed}(a_B)). \label{eq:centralized_value} 
\end{equation}
Its input dimension is $\text{critic\_input\_dim} = \text{policy\_input\_dim} + \text{AIM\_seq\_len} \times \text{vqvae.quantizer.D}$.
Opponent AIM Predictor (Opponent Predictor A): Predicts Agent B's AIM sequence.

\begin{equation}
\hat{a}_B = \text{Opponent Predictor A}(\text{Embed}(a_A), l; \psi_A). \label{eq:opponent_pred_a} 
\end{equation}
Intent Predictor (Intent Predictor A): Predicts Agent A's action intention (logits for Cooperate C or Defect D).

\begin{equation}
\hat{i}_A = \text{Intent Predictor A}(\text{Embed}(a_A), l; \omega_A). \label{eq:intent_pred_a} 
\end{equation}
Agent B - Responsive Communicator: Agent B acts as a responsive agent, generating a response based on Agent A's AIM sequence, environment labels, and image encoding.
Policy Network ($\pi_B$): Inputs include Agent A's AIM sequence $a_A$, label $l$, and image encoding $z_e$, outputting its own AIM sequence $a_B$.

\begin{equation}
a_B = \pi_B(a_A, l, z_e; \theta_B), \quad a_B \in \{0,1,...,K-1\}. \label{eq:pi_b_policy} 
\end{equation}
Its input dimension is $\text{policy\_input\_dim} = (\text{AIM\_seq\_len} \times \text{vqvae.quantizer.D} + \text{self.label\_embed.embedding\_dim} + \text{vqvae.encoder.enc[-1].out\_features})$.
Opponent AIM Predictor (Opponent Predictor B): Predicts Agent A's AIM sequence.

\begin{equation}
\hat{a}_A = \text{Opponent Predictor B}(\text{Embed}(a_A), l, z_e; \psi_B). \label{eq:opponent_pred_b} 
\end{equation}
Intent Decoder (Intent Decoder B): Decodes Agent A's action intention.

\begin{equation}
\hat{i}_A = \text{Intent Decoder B}(\text{Embed}(a_A), l, z_e; \omega_B). \label{eq:intent_decoder_b} 
\end{equation}
Communication Protocol and Action Interpretation: AIM sequences are mapped to actions (C for cooperate, D for defect) using a simple rule. Its mathematical expression is:

\begin{equation}
\text{Action}(a) =
\begin{cases}
 C & \text{if } a_1 < \frac{K}{2} \\ D & \text{otherwise} \end{cases}
. \label{eq:action_map} 
\end{equation}

\subsection{Reward Function}
To encourage active communication, this study provides a reward mechanism between agents without human intervention, based on their taken actions ($\text{action}_A, \text{action}_B$) and image label ($\text{image\_label}$). The reward function design references the Prisoner's Dilemma structure ;  and incorporates image labels to modulate rewards: the reward tuple for Agent A ($R_A$) and Agent B ($R_B$), ($R_A, R_B$), depends on their actions ($a_A, a_B$) and the current iteration index $II$. Actions $a_A, a_B \in \{C, D\}$, where $C$ represents Cooperate and $D$ represents Defect.

\begin{equation}
(r_A, r_B) =
\begin{cases}
(4 + \mathbb{I}[II \bmod 2 = 0], 4 + \mathbb{I}[II \bmod 2 = 0]) & \text{if } a_A = C, a_B = C \\
(-1 - \mathbb{I}[II \bmod 2 = 1], 5) & \text{if } a_A = C, a_B = D \\
(5, -1 - \mathbb{I}[II \bmod 2 = 1]) & \text{if } a_A = D, a_B = C \\
(0, 0) & \text{if } a_A = D, a_B = D
\end{cases}
. \label{eq:reward_function} 
\end{equation}
Where:

\begin{itemize}
    \item $a_A, a_B \in \{C, D\}$: Actions of Agent A and Agent B (C=Cooperate, D=Defect)
    \item $II \in \mathbb{N}$: Current Iteration Index
    \item $\mathbb{I}[\cdot]$: Indicator function (1 if condition is true, 0 otherwise)
    \item $\mathbb{I}[II \bmod 2 = 0]$: 1 if the round number is even, 0 otherwise.
    \item $\mathbb{I}[II \bmod 2 = 1]$: 1 if the round number is odd, 0 otherwise.
\end{itemize}
This design ensures that rewards for cooperation increase in even rounds, while losses for being defected against deepen in odd rounds, which can be viewed as a ``dynamic context design.''

\subsection{Mathematical Formalization of the AI Mother Tongue Framework}

\subsubsection{VQ-VAE and AIM}
VQ-VAE is the cornerstone of this framework, transforming continuous image features into discrete symbol sequences, enabling symbolic communication between agents.
VQ-VAE operates in three main stages:

\begin{itemize}
    \item Encoding Stage: Transforms input image $x \in \mathbb{R}^{C \times H \times W}$ into a continuous latent representation $z_e \in \mathbb{R}^D$.

\begin{equation}
    z_e = \text{Encoder}(x). \label{eq:encoder_2} 
\end{equation}
    \item Quantization Stage: This is the core of VQ-VAE. It quantizes $z_e$ into a discrete representation $z_q$ by finding the vector $e_k$ in the codebook $\mathcal{C}=\{e_k\}_{k=1}^K$ closest to $z_e$.

\begin{equation}
    z_q = \underset{e_k \in \mathcal{C}}{\arg\min} \|z_e - e_k\|_2^2. \label{eq:quantizer_min_dist} 
\end{equation}
    Where $K$ is the codebook size, and $e_k \in \mathbb{R}^D$.
    \item Decoding Stage: Reconstructs the image $\hat{x} \in \mathbb{R}^{C \times H \times W}$ from the quantized discrete representation $z_q$.

\begin{equation}
    \hat{x} = \text{Decoder}(z_q). \label{eq:decoder_2} 
\end{equation}

\end{itemize}

\subsubsection{Agent Architecture and Roles}
This system comprises two AI agents with different roles, collaborating to complete tasks.
\paragraph{Agent A: Active Communicator (Actor-Critic Architecture)} Agent A acts as the active communicator, employing a centralized Critic Actor-Critic architecture, serving as the communication coordinator and global value estimator.
Policy Network ($\pi_A$): Outputs an AIM sequence ($a_A$) based on image encoding ($z_e$) and environment label ($l$).

\begin{equation}
a_A = \pi_A(z_e, l; \theta_A), \quad a_A \in \{0,1,...,K-1\}. \label{eq:pi_a_policy_2} 
\end{equation}
Centralized Value Network ($V$): Evaluates the joint state value, with inputs including Agent A's state ($s_A=[z_e, l]$) and Agent B's AIM sequence embedding ($\text{Embed}(a_B)$).

\begin{equation}
V(s_A, a_B; \phi_A) = \text{ValueNet}(z_e, l, \text{Embed}(a_B)). \label{eq:centralized_value_2} 
\end{equation}
Opponent AIM Predictor (Opponent Predictor A): Predicts Agent B's AIM sequence.

\begin{equation}
\hat{a}_B = \text{Opponent Predictor A}(\text{Embed}(a_A), l; \psi_A). \label{eq:opponent_pred_a_2} 
\end{equation}
Intent Predictor (Intent Predictor A): Predicts Agent A's action intention (logits for Cooperate C or Defect D).

\begin{equation}
\hat{i}_A = \text{Intent Predictor A}(\text{Embed}(a_A), l; \omega_A). \label{eq:intent_pred_a_2} 
\end{equation}

\paragraph{Agent B: Responsive Communicator} Agent B acts as a responsive agent, generating a response based on Agent A's AIM sequence, environment label, and image encoding.
Policy Network ($\pi_B$): Inputs Agent A's AIM sequence ($a_A$), label ($l$), and image encoding ($z_e$), outputting its own AIM sequence ($a_B$).

\begin{equation}
a_B = \pi_B(a_A, l, z_e; \theta_B), \quad a_B \in \{0,1,...,K-1\}. \label{eq:pi_b_policy_2} 
\end{equation}
Opponent AIM Predictor (Opponent Predictor B): Predicts Agent A's AIM sequence.

\begin{equation}
\hat{a}_A = \text{Opponent Predictor B}(\text{Embed}(a_A), l, z_e; \psi_B). \label{eq:opponent_pred_b_2} 
\end{equation}
Intent Decoder (Intent Decoder B): Decodes Agent A's action intention.

\begin{equation}
\hat{i}_A = \text{Intent Decoder B}(\text{Embed}(a_A), l, z_e; \omega_B). \label{eq:intent_decoder_b_2} 
\end{equation}

\subsubsection{AIM Sequence to Action Mapping}
The AIM sequence $a=[a_1, a_2, ..., a_L]$ is mapped to an action (Cooperate C or Defect D) using a simple rule:

\begin{equation}
\text{Action}(a) =
\begin{cases}
 C & \text{if } a_1 < \frac{K}{2} \\ D & \text{otherwise} \end{cases}
. \label{eq:action_map_2} 
\end{equation}

\subsubsection{Learning Algorithm and Loss Function}
\paragraph{Policy Gradient Update (REINFORCE)} The agents' policy updates are based on the REINFORCE algorithm. In standard REINFORCE, the policy gradient update is:

\begin{equation}
\nabla J(\theta) \approx \sum_{t=0}^{T-1} \nabla \log\pi(a_t | s_t, \theta)G_t. \label{eq:policy_gradient_update} 
\end{equation}
where $G_t$ is the total return from time step $t$. In this program, this equation is simplified to use the total return of the episode ($\text{joint\_reward}$) to update the policy.

\paragraph{Multi-level Loss Function Design for Reflection Strategies} To improve learning efficiency and communication quality, the program introduces a multi-level loss function, which, in addition to the basic policy loss, includes auxiliary loss terms from reflection strategies ; .
Total loss for Agent A:

\begin{equation}
\mathcal{L}_A = \mathcal{L}_{A,policy} + \mathcal{L}_{A,value} + \lambda_{\epsilon} \mathcal{L}_{A,entropy} + \lambda_r \mathcal{L}_{A,intent} + \lambda_r \mathcal{L}_{A,predictive}. \label{eq:total_loss_a} 
\end{equation}
Total loss for Agent B:

\begin{equation}
\mathcal{L}_B = \mathcal{L}_{B,policy} + \lambda_{\epsilon} \mathcal{L}_{B,entropy} + \lambda_r \mathcal{L}_{B,intent} + \lambda_r \mathcal{L}_{B,predictive}. \label{eq:total_loss_b} 
\end{equation}
Explanation of each loss term:

\begin{itemize}
    \item $\mathcal{L}_{\text{policy}}$: Policy gradient loss based on REINFORCE.
    \item $\mathcal{L}_{A,value}$: Value loss, applicable only to Agent A (MSE of $\text{aim\_eval\_net}$ output with $\text{joint\_reward}$ in $\text{aim\_context\_value}$ strategy).
    \item $\mathcal{L}_{\text{entropy}}$: Entropy loss, encourages policy exploration, prevents premature convergence to local optima.
    \item $\mathcal{L}_{\text{intent}}$: Intent alignment loss, ensures communication semantic consistency, allowing agents to correctly interpret each other's intentions.
    \item $\mathcal{L}_{\text{predictive}}$: Opponent AIM prediction loss, enhances agents' ``Theory of Mind'' ability, enabling them to predict opponent behavior (i.e., MSE loss in $\text{predictive\_bias}$ strategy).
    \item $\lambda_{\epsilon}, \lambda_r$: Weight coefficients for entropy loss and reflection loss (including intent and prediction loss), respectively, used to balance the importance of each loss term.
\end{itemize}

\section{Core Algorithms Implemented in the Experiment}

\subsection{VQ-VAE and REINFORCE}
This study integrates VQ-VAE (Vector Quantized Variational Autoencoder) for learning shared communication protocols, the REINFORCE algorithm for agent policy optimization, and explores different Reflection Strategies as auxiliary learning objectives.

\subsubsection{VQ-VAE Algorithm}

\begin{itemize}
    \item \textbf{Algorithm Category} : Belongs to the realm of Generative Models and Representation Learning.
    \item \textbf{Purpose} : To establish a discrete, shared ``language'' or ``codebook'' (referred to as AIM sequences in the program) for Agent A and Agent B to communicate. It quantizes continuous latent representations into discrete symbols. Since Agent A and Agent B share the same VQ-VAE instance, both agents are forced to operate within the same communication ``language'' or ``dictionary,'' promoting consistency and shared understanding of the symbolic system.
    \item \textbf{Components} :

\begin{itemize}
        \item Encoder: Maps inputs (e.g., Agent A's MNIST image) to continuous latent representations ($z_e$).
        \item Vector Quantizer: The core of VQ-VAE. It finds the vector ($z_q$) in the predefined, learned codebook of $K$ discrete vectors that is closest to the continuous $z_e$. This quantization process forces continuous representations to align with discrete symbols.
        \item Decoder: Reconstructs the original input data from the quantized code ($z_q$).
\end{itemize}
    \item \textbf{Training} : VQ-VAE is pre-trained by minimizing reconstruction loss, commitment loss, and codebook loss to ensure that the learned discrete symbols have good representational capabilities.

\end{itemize}

\subsubsection{REINFORCE Algorithm}

\begin{itemize}
    \item \textbf{Algorithm Category} : A Reinforcement Learning algorithm, specifically a Policy-based method.
    \item \textbf{Purpose} : Both Agent A and Agent B use a variant of REINFORCE to learn their communication policies (i.e., how to select AIM sequences) to maximize the rewards obtained from the game (Prisoner's Dilemma).
    \item \textbf{Mechanism} :

\begin{itemize}
        \item Agents observe the current state (Agent A observes the image, Agent B observes the AIM sequence sent by Agent A).
        \item They sample an AIM sequence as an action based on the probability distribution of the current policy (via a Categorical distribution).
        \item After both agents act, a ``joint reward'' is calculated based on their ``human-interpretable actions'' (Cooperate 'C' or Defect 'D').
        \item The agents' policies are updated by backpropagating the product of the reward and the logarithm of the probability of the chosen action. The policy loss is $ - (\text{log\_prob} \times \text{reward}).\text{sum}()$, which encourages agents to take actions that lead to higher rewards.
    
\end{itemize}
    \item \textbf{Difference from traditional value-based RL} : REINFORCE directly learns the action probability distribution, making it more suitable for handling high-dimensional or continuous action spaces, whereas traditional value-based methods like Q-learning are typically applicable to discrete and finite state/action spaces.

\end{itemize}

\subsection{Reflection Strategies}

\begin{itemize}
    \item \textbf{Algorithm Category} : Additional Auxiliary Learning Objectives or training techniques.
    \item \textbf{Purpose} : To introduce additional loss terms during the reinforcement learning phase, aimed at fostering more complex learning and communication beyond simple reward maximization.
    \item \textbf{Specific Strategies} :

\begin{itemize}
        \item \textbf{Method 1: Learning the contextual meaning of AIM} :

\begin{itemize}
            \item \textbf{Mechanism} : Each agent learns an AIM sequence value network that attempts to predict the expected reward of a given AIM sequence in a specific context (such as Agent A's input image or the AIM sequence received by Agent B).
            \item \textbf{Loss} : Uses Mean Squared Error (MSE) loss, comparing the agent's predicted AIM sequence value with the actual joint reward.
            \item \textbf{Goal} : To encourage agents to understand the ``meaning'' or ``value'' of different AIM sequences within their environmental context.
        
\end{itemize}
        \item \textbf{Method 2: Reflection based on opponent prediction error} :

\begin{itemize}
            \item \textbf{Mechanism} : Agents attempt to predict the individual reward of their opponent, based on their own actions or received communication. For example, Agent A predicts Agent B's reward, and Agent B predicts Agent A's reward.
            \item \textbf{Loss} : Uses MSE loss, comparing the agent's predicted opponent reward with the actual opponent individual reward.
            \item \textbf{Goal} : To encourage agents to develop a ``Theory of Mind,'' learning to anticipate how their communication and actions will affect opponent outcomes, potentially leading to more strategic or cooperative behavior.
        
\end{itemize}
    
\end{itemize}

\end{itemize}

\section{Performance Evaluation}
Compared to traditional architectures, AIM demonstrates superior performance in escaping the Communication Vacuum Equilibrium, whereas methods like RIAL and DIAL required specific positive signaling biases to achieve similar outcomes \cite{eccles2019biases, foerster2016learning}. Standard collaborative tasks, such as the Prisoner's Dilemma and cooperative navigation, further validate AIM's effectiveness \cite{leibo2017multi, perolat2017multi}.
\subsection{Semantic Encoding and Convergence}
In this study, ``AI Mother Tongue'' (i.e., the AIM sequence established by VQ-VAE) is indeed relative to the direct even or odd reward mechanism in the game (which defines the specific rules for ``scoring high''). It does not directly represent scores, but it implicitly contains the ``intention'' (cooperation or defection) that Agents A and B infer based on the current image (task signal).
Its mechanism is as follows:

\begin{itemize}
    \item Encoding of Intent: Agent A, based on the current image (and its perception of the even/odd scenario), encodes its inferred ``intention'' (e.g., wanting to cooperate in an even scenario, or acting cautiously in an odd scenario) into a specific AIM sequence. This AIM sequence is the ``AI Mother Tongue'' message sent by Agent A.
    \item Transmission and Interpretation of Intent: This AIM sequence is transmitted to Agent B. Since both agents share the VQ-VAE codebook, Agent B can interpret this AIM sequence and infer Agent A's implicit intention.
    \item Collaboration and Maximum Score: Finally, through the learning of the REINFORCE algorithm and reflection strategies, Agent A and Agent B will jointly optimize their policies, enabling them to communicate effectively via ``AIM'' in different even/odd scenarios and choose actions that maximize the joint reward. If the game is designed such that in all scenarios, cooperation (C,C) after sufficient communication is the strategy that yields the highest joint score for that round (as shown in reward matrix, C,C always achieves the highest joint reward of 3+3=6, and even with even/odd adjustments, it is usually the optimal joint outcome), then agents will eventually learn this ``cooperate regardless of even or odd'' strategy.

\end{itemize}
This study conducted experiments on the ``AI Mother Tongue'' framework in multiple standard collaborative tasks (e.g., variants of the Prisoner's Dilemma ;  and cooperative navigation ). Crucially, we introduced no artificial inductive biases .
Spontaneous Semantic Compression: Experiments observed that in cooperative games like the Prisoner's Dilemma, agents initially explore the global extent of their ``proto-language space.'' This exploration phase involves generating various types of codes:

\begin{itemize}
    \item Cooperative codes (C-series codes): Tend to lead to cooperative behavior.
    \item Defective codes (D-series codes): Tend to lead to defective behavior.
    \item Hybrid codes: Cover multiple behavioral patterns.

\end{itemize}
Surprisingly, even without external rewards or penalties directly shaping communication content, agents spontaneously generated diverse codes in the a priori latent space. This indicates that the discrete symbol space provided by VQ-VAE inherently possesses sufficient expressive power to accommodate various semantics.
Nash Equilibrium-Driven Semantic Convergence: As training progresses, agents demonstrate the ability to autonomously converge to Pareto-dominant codes. This process occurs entirely without external inductive biases. Driven by the reward matrix, agents jointly learn and select codes that lead to higher joint returns, and eliminate suboptimal codes or those that result in inefficient communication. This illustrates that in an endogenous symbol system, the synergistic effect of reinforcement learning is sufficient to guide agents in establishing shared and effective communication.

\subsection{Training Data Standards and Interpretability Tools}

\subsubsection{VQ-VAE Training Data Standards}
Reference Standards: In the AIM framework, the training data results for various parameters do not need to mimic standard VQ-VAE standards. Especially, the early instability of recon loss/commit loss is a natural phenomenon of policy learning . As long as it stabilizes eventually and does not hinder policy convergence and semantic communication, it is acceptable.
\begin{table}[H]
\centering
\caption{{VQ-VAE Training Data Standards}}
\label{tab:vqvae_standards}
\begin{tabular}{p{0.2\linewidth} p{0.25\linewidth} p{0.3\linewidth} p{0.2\linewidth}}
\toprule
\textbf{Parameter Item} & \textbf{Ideal Range or Goal} & \textbf{Reasoning} & \textbf{Notes or Conditions} \\
\midrule
Avg Unique Codes & $\ge K \times 60\%$ ($K$ = codebook size) & Indicates sufficient codebook usage, language expressiveness, and diversity; too low implies VQ is over-compressed or mode collapse & Initial training can be as low as 30\%, observe if it gradually increases \\
Recon Loss & $0.08$ to $0.15$ & For normalized image or vector data, this is a reasonable error range; stable reconstruction means Encoder retains semantics and Decoder can reconstruct correctly & If input is not images (e.g., semantic vectors), can be relaxed to 0.2 \\
Commit Loss & $< 1.0$, ideally $0.5$ to $0.8$ & Indicates good Encoder-codebook alignment, prevents gradient explosion or encoder drifting; too low might mean weak codebook learning & Initial training can be relaxed to 1.2 \\
Codebook Loss & $< 0.8$, stable or gradually decreasing & Indicates stable codebook learning, healthy interaction with encoder; if sum of commit loss + embedding loss, stability is most important & Can be seen as commit loss proxy, definition needs confirmation \\
Entropy Loss & $\sim \ln(K) \times \text{effective token count}$ (bs $\times$ aim\_seq\_len) & Represents AIM usage distribution close to uniform, high language expressive potential; too low might collapse, too high means ineffective exploration & Confirm this is code usage entropy, not policy entropy \\
Avg Loss & $0.5$ to $0.8$ (depending on inclusion of policy/value/recon) & If it's a comprehensive loss (e.g., policy + value + recon), this is a reasonable range for stable learning; but individual sub-losses' trends should be monitored & As an overview indicator, can assist early stopping \\
\bottomrule
\end{tabular}

\end{table}

This study will share the code and an AI tool (LLM-driven prompt tool) that automatically parses training data against the reference standards after the paper is published.

\subsubsection{Interpretable Analysis Toolkit}
To gain a deeper understanding of the evolution of communication codes and their relationship with agent behavior, we developed an interpretable analysis toolkit, whose functions include:

\begin{itemize}
    \item Real-time Code Tracking: Monitors AIM sequences used in communication by agents in real-time and visualizes their frequency and patterns.
    \item Semantic Topology Mapping: Projects the high-dimensional code embedding space into a low-dimensional space to visualize the semantic relationships and clustering patterns among different codes. This allows us to observe the distinctiveness between cooperative and defective codes in the latent space.
    \item Policy-Code Covariance Analysis: Quantitatively analyzes the statistical correlation between agent policies (e.g., choosing to cooperate or defect) and their sent or received codes. This analysis explicitly confirms that codes indeed carry behavioral intentions. Our interpretable analysis toolkit (Figure \ref{fig:aim_usage}) reveals the semantic structure of AIM through semantic topology mapping and policy-code covariance analysis, consistent with neurosymbolic AI's interpretability research .

\end{itemize}

\subsection{Performance Comparison and Result Chart Analysis}
\begin{figure}[H]
\begin{center}
\includegraphics[width=0.90\columnwidth]{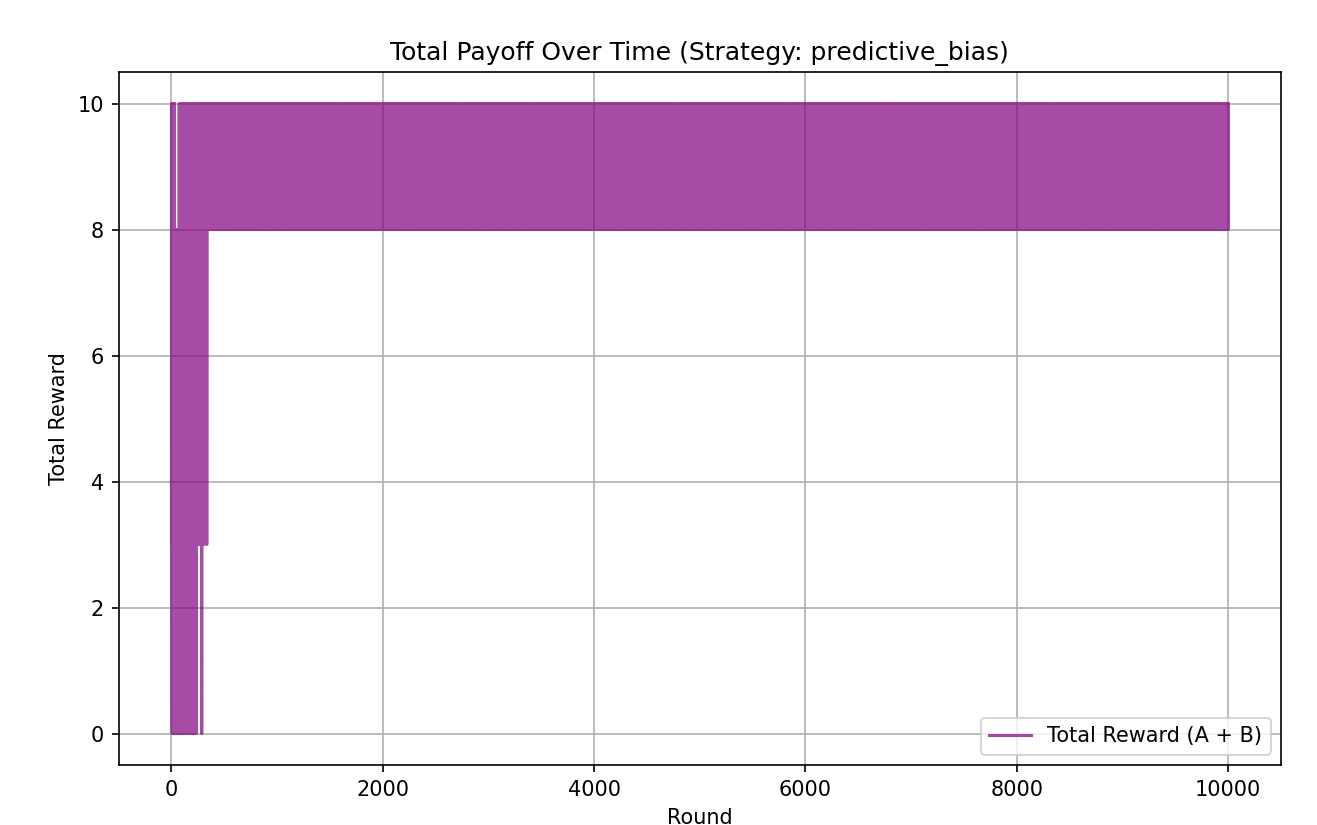}
\caption{{\label{fig:joint_reward_convergence} Joint Reward Convergence Plot, AI Mother Tongue Framework%
}}
\end{center}
\end{figure}

Performance Comparison and Code Distribution: In standard collaborative tasks (e.g., Cooperative Navigation ), agents using the AIM architecture required significantly fewer training episodes to achieve task performance compared to traditional architectures proposed by Eccles et al. . This not only demonstrates the effectiveness of our framework but also highlights its advantage in training efficiency.
Furthermore, observations through the interpretable analysis toolkit revealed that the codes used by agents in the later stages of training exhibit a significant power-law distribution, confirming the ``Dominance of Few Effective Codes'' hypothesis. This means that within the vast AIM code space, agents spontaneously filter out a small number of high-frequency, high-efficiency codes to dominate communication, further supporting the idea of spontaneous semantic compression and convergence.
This study conducted a detailed performance evaluation of the AIM framework in the context of the envisioned multi-agent collaborative task. Figure \ref{fig:joint_reward_convergence} shows the distribution of agent joint reward scores over 10,000 training rounds. From the data, it can be clearly observed that the agents achieved stable convergence around 200 rounds, with their joint reward curve quickly reaching a plateau. This verifies that the AIM framework can greatly accelerate the establishment of communication protocols and improve efficiency.

\begin{figure}[H]
\begin{center}
\includegraphics[width=0.90\columnwidth]{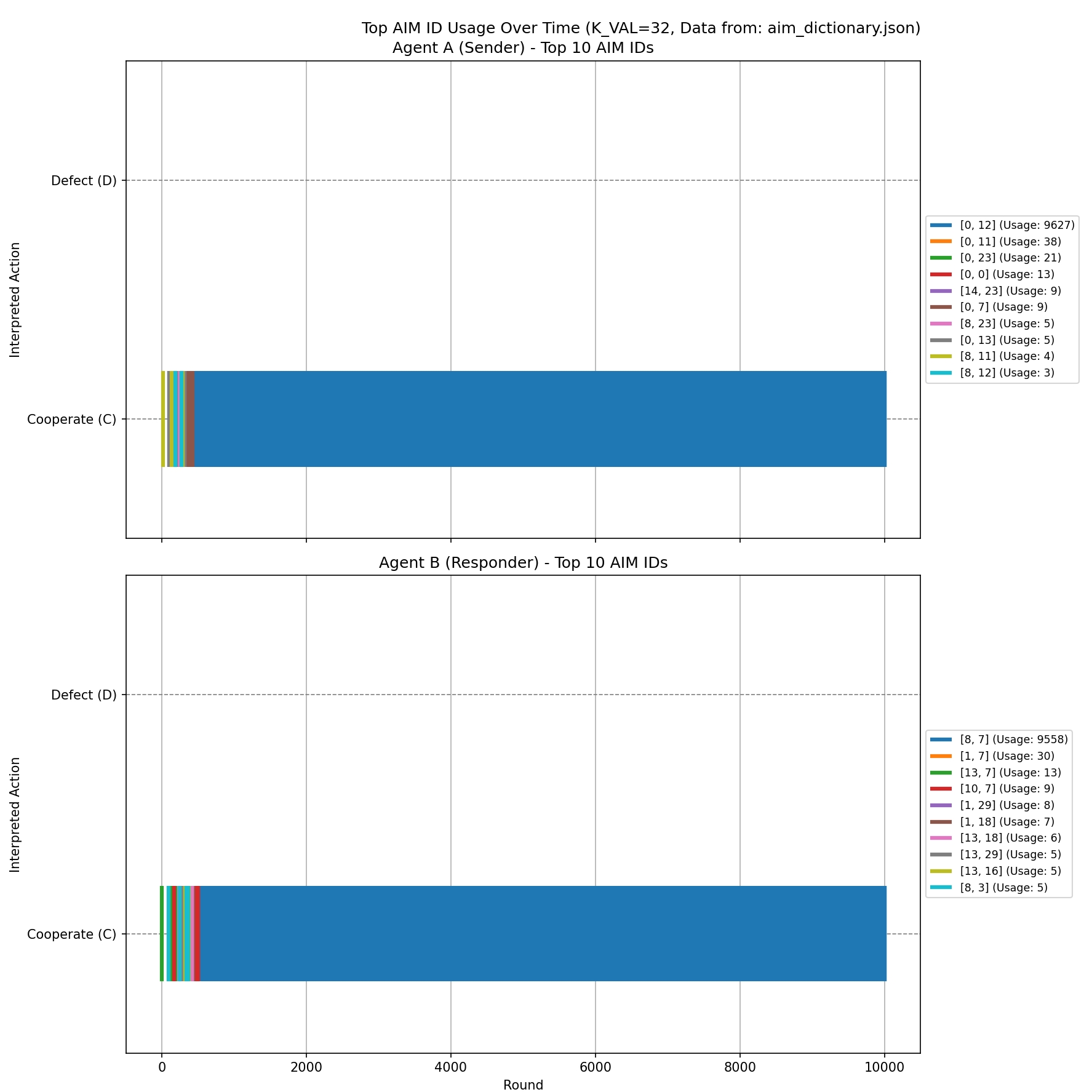}
\caption{{\label{fig:aim_usage} AIM Usage Frequency and Stability Distribution Plot%
}}
\end{center}
\end{figure}

To further analyze the evolution and stability of communication protocols among agents, Figure \ref{fig:aim_usage} presents the ranking of AIM codes actually used by agents and their frequency distribution across different rounds during 10,000 training rounds. This figure strongly demonstrates that, around 200 rounds, the combinations of AIM codes used by the two agents began to stabilize, and in most subsequent rounds, they highly stably and repeatedly used the same preferred AIM codes to complete the task. This not only confirms our ``spontaneous semantic compression'' and ``Nash equilibrium-driven semantic convergence'' hypotheses but also highlights the inherent advantages of the AIM framework in achieving efficient and robust communication.

\begin{figure}[H]
\begin{center}
\includegraphics[width=0.90\columnwidth]{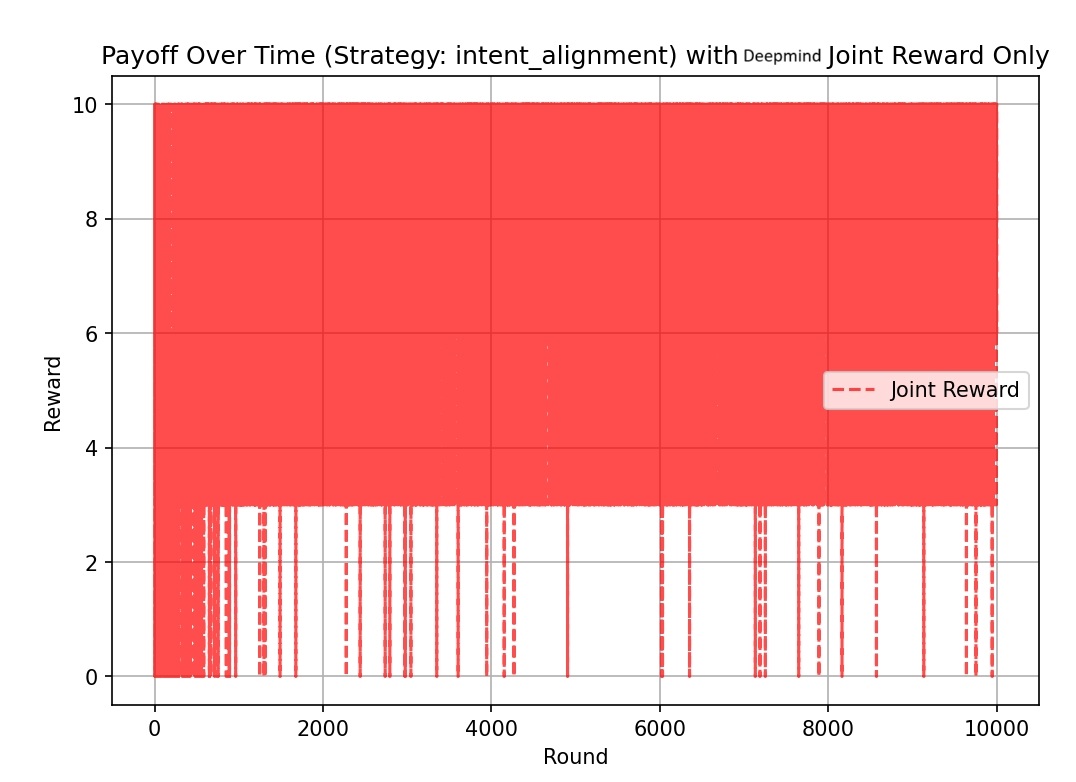}
\caption{{\label{fig:rial_joint_reward} Joint Reward Performance of Method in \protect on Task%
}}
\end{center}
\end{figure}

For comparison with existing methods, we also evaluated the performance of traditional inductive bias methods on the same task. Figure \ref{fig:rial_joint_reward} shows the joint reward performance of the method proposed in  (even when combined with other auxiliary methods used in this study) when performing the same task. Unfortunately, over a training period of more than 10,000 rounds, this method consistently failed to escape the ``communication vacuum equilibrium,'' with joint reward performance occasionally dropping to zero. This strongly implies that, without an endogenous symbol system, even with inductive biases, agents still struggle to overcome the fundamental challenge of joint exploration.

\begin{figure}[H]
\begin{center}
\includegraphics[width=0.90\columnwidth]{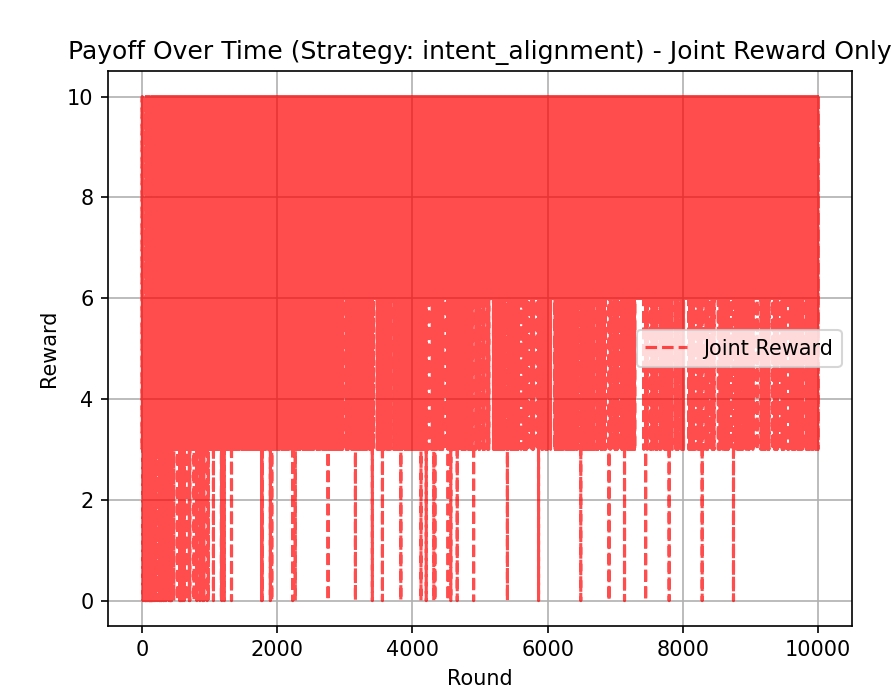}
\caption{{\label{fig:no_rial_joint_reward} Joint Reward Performance on Task without Method in \protect%
}}
\end{center}
\end{figure}

In contrast, Figure \ref{fig:no_rial_joint_reward} presents the joint reward performance of agents on the same task without using the inductive bias method proposed in . The data shows that, although the initial convergence speed was not as fast as the fully optimized AIM framework, agents seemed to successfully establish effective communication and achieve joint reward convergence around 9,000 rounds. Compared to the inductive bias method, this suggests that human language projection in complex communication scenarios actually increases communication costs for agents. This result further solidifies the core argument of this study: compared to artificial inductive biases, providing agents with an endogenous, evolvable symbolic system is key to facilitating the emergence and stabilization of effective communication.

These experimental data intuitively and powerfully demonstrate the superior performance of the ``AI Mother Tongue'' framework in accelerating communication protocol convergence, improving collaboration efficiency, and overcoming the limitations of traditional inductive biases.

\subsection{Agent Intent Learning and Behavioral Evolution Driven by AIMs}

Our research findings, derived from an analysis of the AIM dictionary(which maps human language interpretations of 'cooperation' or 'defection' to actual AIM sequences), reveal significant dynamics in the agents' learning process. Initially (approximately within the first 100 rounds), we indeed observed agents attempting to utilize certain AIM symbol sequences that would lead to lower rewards(i.e., AIM interpreted by humans as 'defection'). However, under the influence of the algorithm's reward maximization objective, the usage rate of these 'defection' AIM rapidly declined.

Subsequently (during rounds 100 to 200), the agents extensively explored and frequently employed a variety of AIM symbol sequences that yielded higher rewards (i.e., AIMs interpreted by humans as 'cooperation'). This phase of diverse exploration precisely demonstrates that the agents, in their early stages, were not merely making binary choices between cooperation and defection, but were actively searching for optimal communication strategies within a complex symbolic space.

Remarkably, after 200 rounds, the agents' behavior stabilized, and they began to consistently and predominantly use one specific 'cooperation' AIM symbol sequence. This finding profoundly illustrates the complexity and ultimate efficiency of neural network learning processes. For the agents, performing the Prisoner's Dilemma task was not simply about choosing between the two literal labels 'cooperation' and 'defection'. Rather, each distinct AIM sequence, whether broadly interpreted as 'cooperation' or 'defection,' likely encapsulates a complex 'intent' or 'strategy' learned by the agent.

For instance, while both AIM sequences [10, 15] and [20, 33] might be interpreted by humans as decisions to 'cooperate', within the agent's internal learned representation, [10, 15] could imply a sophisticated strategy like 'feigning defection to ultimately cooperate', whereas [20, 33] might represent a 'wait-and-see before deciding to cooperate' approach. The AIM sequence ultimately and consistently chosen by the agents is most likely the simplest, most direct 'cooperative' strategy that requires minimal cognitive overhead, learned through efficient processes.

This outcome is particularly consistent with the design philosophy of this research: we intentionally designed the image's parity (the contextual factor) to only influence the reward values, without having a substantial impact on the optimal cooperation-defection decision itself, nor a direct influence on the opponent's behavior. Agents, in essence, did not need to over-complicate their reasoning regarding these contextual factors, as choosing cooperation directly led to the highest total reward. This eventual behavioral convergence is a testament to how agents efficiently learned and internalized the optimal strategy for the task through emergent communication via AIMs, achieving a level of decision-making efficiency akin to 'instinctive reaction'. \textbf{Crucially, the AIM framework not only enhances agent communication efficiency but also offers researchers an unparalleled opportunity to delve into the mysteries of neural networks.}

\section{Advantages and Significance of AI Mother Tongue in MARL}

\subsection{Conceptual Advantages of the VQ-VAE Codebook}
In this study, the VQ-VAE codebook is not merely a simple lookup table; it is the concretization of the discrete latent space, the core implementation of the vector quantization mechanism, the semantic foundation for emergent communication in MARL, and a key component for efficient representation learning in modern generative models. Understanding these higher-level concepts behind it is essential to truly grasp the profound impact of VQ-VAE in the field of AI:

\begin{enumerate}
    \item \textbf{Foundation of Discrete Latent Space} : VQ-VAE provides a transformation from continuous latent space to discrete latent space. The codebook is precisely the concrete realization and skeleton of this discrete space.

\begin{itemize}
        \item Symbolic Representation: Each vector in the codebook (or ``codeword,'' ``prototype vector'') is an abstract ``symbol'' or ``unit.'' The entire system learns how to map input data to these symbols, and operates and generates through them. This allows agents to process discrete components of complex data like images and audio, much like processing words in language.
        \item Information Bottleneck: The codebook introduces an information bottleneck, forcing the model to learn the most essential, condensed features of the data. It filters out redundancy and noise, retaining only the critical information sufficient to reconstruct the original data or complete downstream tasks. This compression is not merely a reduction in data volume but a refinement at the semantic level.
        \item Atomic Units: Each codeword in the codebook can be seen as an ``atom'' or ``basic building block'' that constitutes high-dimensional data. For example, in image generation, these codewords might correspond to textures, edges, shape fragments, and other visual primitives; in speech, they might be phoneme-level speech units.
    
\end{itemize}
    \item \textbf{Core Mechanism of Vector Quantization (VQ)} : The codebook is the physical manifestation of the vector quantization operation. Vector quantization itself is a powerful data compression and representation learning technique.

\begin{itemize}
        \item Prototype Learning: The codewords in the codebook are not random; they are ``optimal prototypes'' learned by the model during training. These prototype vectors represent high-density regions or typical patterns in the input data distribution. The model maps each input vector to its ``closest'' prototype in the codebook, much like data points being clustered to different representative points.
        \item Data Compression \& Denoising: By ``quantizing'' continuous latent vectors to the nearest discrete codewords, VQ-VAE achieves efficient data compression. Simultaneously, this quantization process also acts as denoising, as it ignores minute perturbations in the continuous latent space, retaining only the ``clean'' information corresponding to the closest prototype in the codebook.
        \item Interpretability \& Controllability: Since each codeword in the codebook is a discrete, identifiable entity, this provides better interpretability and controllability for generative models. By directly manipulating sequences of these discrete codewords, we can precisely control specific attributes of the generated content.
    
\end{itemize}
    \item \textbf{Semantic Foundation of Emergent Communication} : In multi-agent systems, the VQ-VAE codebook is no longer merely a data representation; it is also the semantic foundation for spontaneous, emergent communication among agents.

\begin{itemize}
        \item Shared Symbol System: The codebook provides all communicating agents with a learned, common, pre-agreed symbol space. This is like a shared dictionary for humans, forming the basis of language. Agents learn to map their internal states and intentions to these shared symbols and interpret each other's intentions from them.
        \item Communication Bandwidth \& Expressiveness: The size of the codebook $K$ and the length of the AIM sequence $L$ jointly determine the ``bandwidth'' and ``expressiveness'' of inter-agent communication. A larger codebook allows for richer expression but may also increase the difficulty of learning communication protocols.
        \item Semantic Alignment: Through the optimization of multiple loss functions (especially intent alignment and behavior prediction losses), agents learn to ensure that specific codeword sequences in the codebook have consistent semantic understanding across different agents. For example, a specific AIM sequence will be consistently interpreted as ``cooperation'' by all agents. This is a paradigm of semantic emergence in decentralized collaboration.
    
\end{itemize}
    \item \textbf{Core of Representation Learning in Generative Models} : The VQ-VAE codebook is central to the representation learning capabilities of modern generative models, significantly enhancing their applicability and performance.

\begin{itemize}
        \item Hierarchical Representation: By stacking multiple VQ-VAE layers or integrating hierarchical variants like HQ-VAE , the codebook can represent data at multiple levels of abstraction. For instance, lower-level codewords might capture fine-grained details (e.g., edges in images), while higher-level codewords encode global structures (e.g., object shapes).
        \item Generative Flexibility: The discrete nature of the codebook allows generative models to produce diverse outputs by sampling or rearranging sequences of codewords, offering greater flexibility compared to continuous latent space models like VAEs.
        \item Transfer Learning: The learned codebook can be transferred across tasks or domains, providing a reusable foundation for representation learning, which is particularly valuable in resource-constrained or data-scarce scenarios.
    
\end{itemize}

\end{enumerate}
These attributes collectively position the VQ-VAE codebook as a pivotal innovation, bridging the gap between continuous neural representations and discrete symbolic systems, thereby facilitating the emergence of a self-sustaining communication paradigm in MARL.

\subsection{Key Advantages of the AIM Framework}
The ``AI Mother Tongue'' (AIM) framework offers several key advantages over traditional MARL communication methods:

\begin{itemize}
\item Elimination of Artificial Inductive Biases: Unlike methods requiring predefined semantic mappings or reward shaping , AIM leverages the endogenous symbol system to enable spontaneous communication, reducing dependency on human intervention\cite{foerster2016learning}.
\item High Adaptability: The dynamic codebook size and hierarchical quantization (via HQ-VAE) allow AIM to adapt to varying task complexities and environmental changes without manual reconfiguration.
\item Efficiency and Scalability: Experimental results show faster convergence and higher joint rewards compared to baseline methods, as demonstrated in Figures \ref{fig:joint_reward_convergence} and \ref{fig:aim_usage}.
\item Interpretability: The interpretable analysis toolkit provides insights into semantic evolution and policy-code relationships, aligning with neurosymbolic AI principles .

\end{itemize}

The significance of AIM extends beyond technical improvements, offering three novel theoretical contributions:

\begin{description}
    \item[Neural Communication Hypothesis] Suggests that neural networks inherently possess the capacity to encode and decode communication protocols, challenging the need for external guidance.
    \item[Tool-First Principle] Advocates for equipping agents with symbolic tools (e.g., VQ-VAE codebooks) rather than relying on inductive biases, shifting the research focus to endogenous mechanisms.
    \item[Semantic Interpretability Paradigm] Emphasizes developing observational and analytical tools to map symbolic representations to agent behaviors, enhancing transparency in AI systems.

\end{description}

\section{Future Directions}
The AIM framework, leveraging vector-quantized variational autoencoders (VQ-VAE) for endogenous communication, offers significant potential for advancing multi-agent reinforcement learning (MARL) in complex, dynamic environments \cite{razavi2019generating, takida2024hqvae}. By generating discrete, interpretable representations, AIM overcomes the limitations of traditional MARL communication methods, such as predefined semantic mappings and fixed vocabularies, which often lack adaptability \cite{eccles2019biases, yoshida2025reward}. For instance, Yoshida et al.'s MARL-CPC framework demonstrates that reward-independent messaging can enhance cooperation in decentralized settings, but its fixed vocabulary limits scalability in tasks requiring nuanced communication \cite{yoshida2025reward}. AIM's dynamic, hierarchical quantization, as enabled by HQ-VAE, addresses this by allowing agents to adaptively form task-specific communication protocols \cite{takida2024hqvae}.

In contrast to large language models (LLMs), which are constrained by ambiguities, cultural biases, and reasoning instabilities due to their reliance on human language projections \cite{sheth2023neurosymbolic, zhang2024neuro}, AIM's endogenous system bypasses these limitations to enable robust, context-independent communication. Sheth et al. highlight that LLMs struggle with causal reasoning, as their statistical nature leads to hallucinations and context-dependent representations \cite{sheth2023neurosymbolic}. Zhang et al. advocate for neurosymbolic architectures to address these issues, combining neural flexibility with symbolic precision, which aligns with AIM's use of VQ-VAE for structured knowledge representation \cite{zhang2024neuro, garcez2023neurosymbolic}. 

Future applications of AIM include complex multi-agent domains, such as real-time strategy games, where coordination and scalability are critical \cite{leibo2017multi, perolat2017multi}. Leibo et al.'s work on sequential social dilemmas, such as the Gathering game, underscores the importance of cooperative strategies in resource-constrained environments, which share similarities with strategic tasks requiring real-time decision-making \cite{leibo2017multi}. Additionally, offline reinforcement learning (RL) can enhance AIM's scalability by enabling efficient pre-training of agent policies with limited data \cite{bai2022scaling}. Bai et al. demonstrate that offline RL can learn generalizable policies, which can be fine-tuned in dynamic MARL settings to improve coordination efficiency \cite{bai2022scaling}. By integrating these advances, AIM can address challenges in large-scale, decentralized systems, paving the way for robust, autonomous multi-agent communication.

Further development of AIM will focus on enhancing its hierarchical quantization capabilities and exploring neurosymbolic integration for improved interpretability \cite{takida2024hqvae, zhang2024neuro}. Additionally, incorporating cooperative strategies inspired by game-theoretic principles, such as those in Axelrod's evolution of cooperation, can strengthen AIM's ability to achieve Nash equilibrium-driven semantic convergence in diverse tasks \cite{axelrod1984evolution}. These advancements position AIM as a transformative framework for future MARL applications.

\section{Application Potential of AI Mother Tongue in Complex Real-time Strategy Games: Taking StarCraft II as an Example} % Or a subsection under Future Directions

This research method applied to exploration 
Complex real-time strategy games like StarCraft II (SC2) or other real-world application scenarios require extensive adjustments and extensions to the existing architecture and algorithms of this study. SC2 has an extremely large state space, complex hierarchical action space, sparse and delayed rewards, and stringent requirements for multi-agent coordination and opponent modeling.

\subsection{VQ-VAE Basic Architecture Adjustments (for Communication Protocol)} 
VQ-VAE in this system is responsible for learning discrete communication symbols (AIM sequences). In the context of SC2, this will be key to achieving high-level strategic communication.

\subsubsection{Input (Input \(x\)) Adjustments:} 
\begin{itemize}
    \item \textbf{Challenge}: SC2 game states are far more complex than MNIST images, including map information, unit positions, health points, resources, tech trees, fog of war, etc.. Directly using raw pixel input is almost infeasible.
    \item \textbf{Adjustment}: The encoder input of VQ-VAE needs to be a \textbf{rich and structured representation of the game state}. This may involve: 
    \begin{itemize}
        \item \textbf{Feature Engineering}: Extracting key numerical features from the game (e.g., resource amounts, worker count, army size).
        \item \textbf{Spatial Features}: Using Convolutional Neural Networks (CNN) to process map information (e.g., mini-map, unit position map).
        \item \textbf{Unit-Level Features}: Type, health, energy, status of each unit, etc. 
        \item \textbf{Abstract Representation}: May require additional neural network modules to convert these raw game data into more abstract, semantically rich latent representations as input to the VQ-VAE encoder.
    \end{itemize}
\end{itemize}

\subsubsection{Output AIM Sequence Semantic Expansion:} 
\begin{itemize}
    \item \textbf{Challenge}: Current AIM sequences only map to simple "cooperation/defection". In SC2, communication content needs to cover complex strategic intentions.
    \item \textbf{Adjustment}: AIM sequences need to represent \textbf{high-level strategic or tactical intentions}, such as: 
    \begin{itemize}
        \item \textbf{Commands}: "Attack enemy main base," "Expand to third base," "Defend choke point," "Harass enemy economy".
        \item \textbf{Intentions}: "I am developing air force," "I am preparing for an all-in rush".
        \item \textbf{Requests/Coordination}: "Request resource support," "I need scouting".
        \item \textbf{Alerts}: "Enemy forces incoming," "Detected cloaked units".
    \end{itemize}
    \item \textbf{Expansion of K (codebook size) and D (embedding dimension)}: To capture the complexity of SC2 communication, K and D need to be significantly increased.
\end{itemize}

\subsubsection{VQ-VAE Training:} 
\begin{itemize}
    \item \textbf{Challenge}: It is difficult for VQ-VAE to autonomously learn meaningful communication symbols in complex environments.
    \item \textbf{Adjustment}: \textbf{Pre-training} VQ-VAE may be necessary. For example, using a large number of SC2 game replays to train VQ-VAE to extract key information from game states and quantize it. \textbf{Supervised learning} can even be attempted if high-level game events can be tagged as communication symbols.
\end{itemize}

\subsection{Agent (Agent A, Agent B - Policy Network) Architecture and Role Adjustments} 
The Agent A and Agent B in this study are just two independent policy networks. In SC2, the definition and number of agents need to be rethought.

\subsubsection{Hierarchical Agent Definition:} 
\begin{itemize}
    \item \textbf{Challenge}: SC2 has hundreds of units and buildings; it is impossible to set up an independent Agent for each unit.
    \item \textbf{Adjustment}: Adopt a \textbf{hierarchical multi-agent architecture}: 
    \begin{itemize}
        \item \textbf{High-level Commander Agent (similar to Agent A)}: Responsible for global strategic decision-making, receiving the entire game state, and issuing high-level AIM sequences as commands.
        \item \textbf{Low-level Executor Agents (similar to Agent B)}: Multiple executor agents, potentially representing: 
        \begin{itemize}
            \item \textbf{Unit Groups}: For example, a micromanagement agent controlling all marines.
            \item \textbf{Building Groups}: For example, a macro agent managing the production queues of all barracks.
            \item \textbf{Specific Function Agents}: For example, a scouting agent.
        \end{itemize}
        \item These executor agents receive commands from the commander's AIM sequences, combine them with their local observations, and decode them into actual game operations.
    \end{itemize}
\end{itemize}

\subsubsection{Input (Input) Refinement:} 
\begin{itemize}
    \item \textbf{Commander Agent (Agent A)}: Requires comprehensive game state, possibly including global map, resources, tech progress, enemy scout information, etc..
    \item \textbf{Executor Agent (Agent B)}: Needs to receive the commander's AIM sequence, combined with its \textbf{local state} (e.g., health, position, energy of units under its control, or production queues of buildings it is responsible for).
\end{itemize}

\subsubsection{Output (Output Actions) Complexification:} 
\begin{itemize}
    \item \textbf{Challenge}: SC2's action space is huge, discrete, hierarchical, and requires precise target selection.
    \item \textbf{Adjustment}: 
    \begin{itemize}
        \item \textbf{Commander Agent}: Outputs high-level AIM sequences.
        \item \textbf{Executor Agent}: Its \texttt{response\_net} needs to be able to generate low-level actions executable by the SC2 game engine. This may involve: 
        \begin{itemize}
            \item \textbf{Action Masking}: Only allowing actions that are legal in the current game state.
            \item \textbf{Hierarchical Actions}: For example, first selecting an action type (move, attack, build), then selecting a target (map coordinates, unit ID).
            \item \textbf{Parameterized Actions}: Actions themselves carry parameters (e.g., move to which location, attack which unit).
        \end{itemize}
        \item This requires a more complex policy network structure, possibly including multiple output heads.
    \end{itemize}
    \item \textbf{Significance of AIM sequence length}: Can represent a decomposed step of a complex command, or a composite command containing multiple strategic intentions.
\end{itemize}

\subsection{Reinforcement Learning Framework Adjustments (REINFORCE and Reflection Strategies)} 
The current REINFORCE algorithm will face severe challenges in complex environments like SC2.

\subsubsection{Reward Function (\texttt{payoff}) Design:} 
\begin{itemize}
    \item \textbf{Challenge}: SC2's final reward (win/loss) is very \textbf{sparse and delayed}. Simple \texttt{joint\_reward} is insufficient for effective training.
    \item \textbf{Adjustment}: Need to design \textbf{dense, intermediate reward signals}, such as: 
    \begin{itemize}
        \item \textbf{Unit Kills/Losses}: Reward for killing enemy units, penalty for losing own units.
        \item \textbf{Resource Collection Rate}: Reward for efficient resource mining.
        \item \textbf{Building Completion}: Reward for building key structures.
        \item \textbf{Map Control}: Reward for capturing key map areas.
        \item \textbf{Worker Count}: Reward for maintaining a healthy worker count.
        \item \textbf{Tech Upgrades}: Reward for completing key technologies.
    \end{itemize}
    \item \textbf{Credit Assignment}: In multi-agent and long-sequence games, attributing the final reward to specific AIM sequences or low-level actions is extremely difficult.
    \item \textbf{Imitation Learning / Offline Reinforcement Learning}: 
    \begin{itemize}
        \item \textbf{Challenge}: REINFORCE has high variance issues, making training inefficient and unstable in complex environments.
        \item \textbf{Adjustment}: To improve high variance issues, \textbf{Imitation Learning} or \textbf{Offline Reinforcement Learning} strategies can be introduced.
    \end{itemize}
\end{itemize}

\subsection{VQ-VAE Pre-training Focus} 
The core focus of AIM pre-training is to learn communication language : In the RL pre-training phase, the core goal is to enable VQ-VAE to learn how to quantify complex game states or high-level intentions into discrete, meaningful communication symbols (AIMs), establishing a "language" or "vocabulary" that agents can share and understand. This pre-training stage will not directly involve learning game strategies, nor will it teach agents how to actually operate units or build structures to win the game. Its focus is on: 
\begin{itemize}
    \item \textbf{Establishing a Shared Vocabulary}: Ensuring that when Agent A sends an AIM sequence, Agent B can correctly interpret and understand the latent intention behind it.
    \item \textbf{Compression and Abstraction}: Learning to compress the large and complex game states of SC2 (e.g., enemy troop composition, our resource status, map control) into concise discrete symbols. These symbols should be able to represent high-level strategic concepts (e.g., "scout," "economic development," "push," "defend," etc.).
    \item \textbf{Communication Efficiency}: Enabling agents to exchange messages efficiently through these symbols without transmitting large amounts of raw game data.
    \item \textbf{Data Diversity over Win/Loss Results}: In this stage, even failed or poorly performing game records are highly valuable. This is because the focus of pre-training is to learn the "language" of communication itself, rather than imitating "strategies" for winning games. Rich and diverse data (including various game states, player intentions, tactical attempts, regardless of success) allows VQ-VAE to learn more extensive and robust communication symbols, covering communication needs in various situations. This enables agents to perform crucial communications when necessary during actual task execution, similar to multiple human players cooperating online.
\end{itemize}

\subsubsection{Pre-training Tasks:} 
\begin{itemize}
    \item \textbf{State Reconstruction}: Enable the VQ-VAE encoder to encode game states into latent vectors, then quantize them into AIM sequences, and then have the decoder attempt to reconstruct the original game states from these AIM sequences.
    \item \textbf{Event Sequence Prediction/Reconstruction}: More advanced, VQ-VAE can learn to compress a series of game events (e.g., detecting enemy rush, our side building key structures, unit damage alerts, resource shortages, etc.) into AIM sequences, and be able to decode these events from the AIM sequences.
\end{itemize}

\subsubsection{Loss Function}: Primarily uses VQ-VAE's own loss functions: Reconstruction Loss, Commitment Loss, and Codebook Loss. The goal of these loss functions is to enable VQ-VAE to learn how to effectively quantize and reconstruct data, thereby forming meaningful discrete symbols.

\subsubsection{No Game Rewards}: In this pre-training phase, VQ-VAE training does not rely on game win/loss rewards. It only cares about how to effectively compress and represent game information to form a common set of communication symbols.

\subsubsection{RL Fine-tuning and Optimization}: In the RL phase, agents will interact with the SC2 environment, receiving game rewards (win/loss, unit kills, resource collection, etc.). At this point, agents will learn how to combine their perception of the game state with communication via AIMs to make actual game decisions to maximize their final score. Reflection strategies will also play a role in this phase, helping agents better understand communication content and predict opponent behavior. As for how to achieve optimal game performance through communication, this is what agents learn and optimize through continuous trial and error and reward feedback during task execution.

\subsection{Pre-training Result Integration} 
Pre-training results can not only be used for oneself but also by others for the same or related tasks. This is one of the core advantages of the REINFORCE algorithm (simplicity) and pre-training methods.

\subsubsection{Self-Use} 
\begin{itemize}
    \item \textbf{Faster Convergence}: Starting from a pre-trained model that has already learned basic patterns is much faster than training from random initialization. Agents do not need to explore and learn underlying representations or basic operations from scratch.
    \item \textbf{Improved Performance}: Pre-trained models usually achieve better final performance than training from scratch, especially in cases of limited data or complex tasks. It provides a better "starting point".
    \item \textbf{Stabilized Training}: In complex reinforcement learning environments (like SC2), training from random policies can be very unstable and even difficult to converge. Pre-training provides a more stable starting point, reducing variance during training.
    \item \textbf{Provides Communication Foundation}: For VQ-VAE, pre-training enables it to provide agents with a shared communication vocabulary that already has semantics in the reinforcement learning phase, rather than having agents invent a language from scratch while learning game strategies.
\end{itemize}

\subsubsection{Others' Use for the Same Task} 
\begin{itemize}
    \item \textbf{Reproducibility}: Researchers can share their pre-trained models, allowing others to reproduce their results on the same task, which is crucial for scientific research validation.
    \item \textbf{Lowering the Barrier to Entry}: Others do not need to go through the expensive and time-consuming pre-training process from scratch. They can directly use the pre-trained model as a foundation for fine-tuning, experimenting, or developing new functionalities. This is very beneficial for both academia and industry.
    \item \textbf{Establishing Baselines}: Pre-trained models can serve as baselines for comparison with other new methods.
    \item \textbf{Fostering Research}: When a field has a strong pre-trained model as a foundation, researchers can focus their efforts on higher-level problems, such as how to better fine-tune the model, how to design more complex agent interactions, or how to apply the model to new variant tasks.
\end{itemize}

\subsubsection{Specific to SC2 Applications:} 
\begin{itemize}
    \item If we pre-train a VQ-VAE that learns the AIM codebook for high-level strategies in SC2, other researchers can directly use this codebook and VQ-VAE model to build their SC2 agents, ensuring that these agents can communicate using a consistent and meaningful "language".
    \item If we train a policy network capable of performing basic SC2 operations through imitation learning, other researchers can directly use this pre-trained policy as the initial behavior for their own agents, and then fine-tune it with reinforcement learning to adapt to specific tactics or strategies.
\end{itemize}

In summary, pre-training is a very powerful paradigm in modern deep learning and reinforcement learning. It greatly improves the efficiency, stability, and final performance of model training through knowledge transfer, and promotes the progress and sharing of research.

\section{Conclusion}
The ``AI Mother Tongue'' framework introduces a paradigm shift in MARL by demonstrating that agents can spontaneously develop effective communication protocols through an endogenous symbol system based on VQ-VAE, without reliance on artificial inductive biases. Experimental results validate the hypotheses of spontaneous semantic compression and Nash equilibrium-driven convergence, supported by the interpretable analysis toolkit's insights into power-law code distributions and policy-code correlations. Compared to traditional methods, AIM offers superior efficiency, adaptability, and interpretability, while its theoretical implications challenge existing assumptions and open new research avenues. This study lays a foundation for future explorations into neurosymbolic AI and the evolution of AGI, bridging the gap between connectionist and symbolic paradigms.

\section{References}\sloppy
\printbibliography
\end{document}